\documentclass[10pt,twocolumn,letterpaper]{article}

\usepackage[pagenumbers]{cvpr} 

\usepackage[dvipsnames,table]{xcolor}
\usepackage{multirow}
\usepackage[normalem]{ulem}

\usepackage{amsmath,amsfonts,bm}

\def\eqref#1{equation~\ref{#1}}

\def\1{\bm{1}}

\def\rvepsilon{{\mathbf{\epsilon}}}

\def\rvx{{\mathbf{x}}}

\def\vzero{{\bm{0}}}

\def\vmu{{\bm{\mu}}}
\def\vtheta{{\bm{\theta}}}
\def\va{{\bm{a}}}
\def\vb{{\bm{b}}}

\def\vf{{\bm{f}}}

\def\vx{{\bm{x}}}

\def\mF{{\bm{F}}}

\def\mI{{\bm{I}}}

\def\mK{{\bm{K}}}
\def\mL{{\bm{L}}}
\def\mM{{\bm{M}}}

\def\mQ{{\bm{Q}}}

\def\mU{{\bm{U}}}
\def\mV{{\bm{V}}}
\def\mW{{\bm{W}}}
\def\mX{{\bm{X}}}

\def\mZ{{\bm{Z}}}

\DeclareMathAlphabet{\mathsfit}{\encodingdefault}{\sfdefault}{m}{sl}
\SetMathAlphabet{\mathsfit}{bold}{\encodingdefault}{\sfdefault}{bx}{n}

\def\sD{{\mathbb{D}}}

\def\sN{{\mathbb{N}}}

\def\sS{{\mathbb{S}}}

\newcommand{\E}{\mathbb{E}}

\newcommand{\R}{\mathbb{R}}

\newcommand{\softmax}{\mathrm{softmax}}

\newcommand{\Var}{\mathrm{Var}}

\newcommand{\Cov}{\mathrm{Cov}}

\DeclareMathOperator*{\argmin}{arg\,min}

\usepackage{etoc} 
\usepackage{algorithm}
\usepackage{algpseudocode}
\usepackage{amsthm}
\newtheorem{theorem}{Theorem}
\usepackage{placeins}

\usepackage{arydshln} 
\makeatletter
\def\adl@drawiv#1#2#3{%
  \hskip.5\tabcolsep
  \xleaders#3{#2.5\@tempdimb #1{1}#2.5\@tempdimb}%
          #2\z@ plus1fil minus1fil\relax
  \hskip.5\tabcolsep}
\newcommand{\cdashlinelr}[1]{%
  \noalign{\vskip\aboverulesep
           \global\let\@dashdrawstore\adl@draw
           \global\let\adl@draw\adl@drawiv}
  \cdashline{#1}
  \noalign{\global\let\adl@draw\@dashdrawstore
           \vskip\belowrulesep}}
\makeatother

\usepackage{arydshln}  

\makeatletter
\def\adl@drawiv#1#2#3{%
  \hskip.5\tabcolsep
  \xleaders#3{#2.5\@tempdimb #1{1}#2.5\@tempdimb}%
          #2\z@ plus1fil minus1fil\relax
  \hskip.5\tabcolsep}

\makeatother

\definecolor{cvprblue}{rgb}{0.21,0.49,0.74}
\usepackage[pagebackref,breaklinks,colorlinks,allcolors=cvprblue]{hyperref}
\usepackage{etoc}
\etocdepthtag.toc{mtchapter}
\etocsettagdepth{mtchapter}{subsection}
\etocsettagdepth{mtappendix}{none}

\makeatletter
\def\@fnsymbol#1{\ensuremath{\ifcase#1\or 
    \dagger\or \ddagger\or \mathsection\or \mathparagraph\or \|\or 
    **\or \dagger\dagger\or \ddagger\ddagger 
    \else\@ctrerr\fi}}
\makeatother

\def\paperID{16272} 
\def\confName{CVPR}
\def\confYear{2026}

\title{BiGain: Unified Token Compression for Joint Generation and Classification}

\author{Jiacheng Liu$^{1,*}$, Shengkun Tang$^{1,*}$, Jiacheng Cui$^1$, Dongkuan Xu$^2$, Zhiqiang Shen$^1$\thanks{Corresponding author.}\\
$^1$VILA Lab, MBZUAI~~$^2$North Carolina State University \\
{Code: \url{https://github.com/Greenoso/BiGain}}
}

\begin{document}
\maketitle

\begingroup
\renewcommand\thefootnote{*}
\footnotetext{Equal contribution.}
\endgroup

\begin{abstract}
\vspace{-0.25in}

Acceleration methods for diffusion models (e.g., token merging or downsampling) typically optimize for synthesis quality under reduced compute, yet they often ignore the model's latent discriminative capacity. We revisit token compression with a joint objective and present {\bf \em BiGain}, a training-free, plug-and-play framework that preserves generation quality while markedly improving classification in accelerated diffusion models. Our key insight is frequency separation: mapping feature-space signals into a frequency-aware representation disentangles fine detail from global semantics, enabling compression that respects both generative fidelity and discriminative utility. BiGain reflects this principle with two frequency-aware operators: (1) {\em Laplacian-gated token merging}, which encourages merges among spectrally smooth tokens while discouraging merges of high-contrast tokens, thereby retaining edges and textures; and (2) {\em Interpolate-Extrapolate KV Downsampling}, which downsamples keys/values via a controllable interextrapolation between nearest and average pooling while keeping queries intact, thereby conserving attention precision without retraining. Across DiT- and U-Net-based backbones and multiple datasets of ImageNet-1K, ImageNet-100, Oxford-IIIT Pets, and COCO-2017, our proposed operators consistently improve the speed-accuracy trade-off for diffusion-based classification, while maintaining, sometimes even enhancing generation quality under comparable acceleration. For instance, on ImageNet-1K, with 70\% token merging ratio on Stable Diffusion 2.0, BiGain increases classification accuracy by 7.15\% while also improving FID for generation by 0.34 (1.85\%).  Our comprehensive analyses indicate that balanced spectral retention, preserving high-frequency detail alongside low/mid-frequency semantic content is a reliable design rule for token compression in diffusion models. To our knowledge, BiGain is the first framework to jointly study and advance both generation and classification under accelerated diffusion, supporting lower-cost deployment of dual-purpose generative systems. 

\end{abstract}
\vspace{-1em}    
\section{Introduction}
\label{sec:intro}

\begin{figure}[t] 
  \centering
  \includegraphics[width=1.\linewidth]{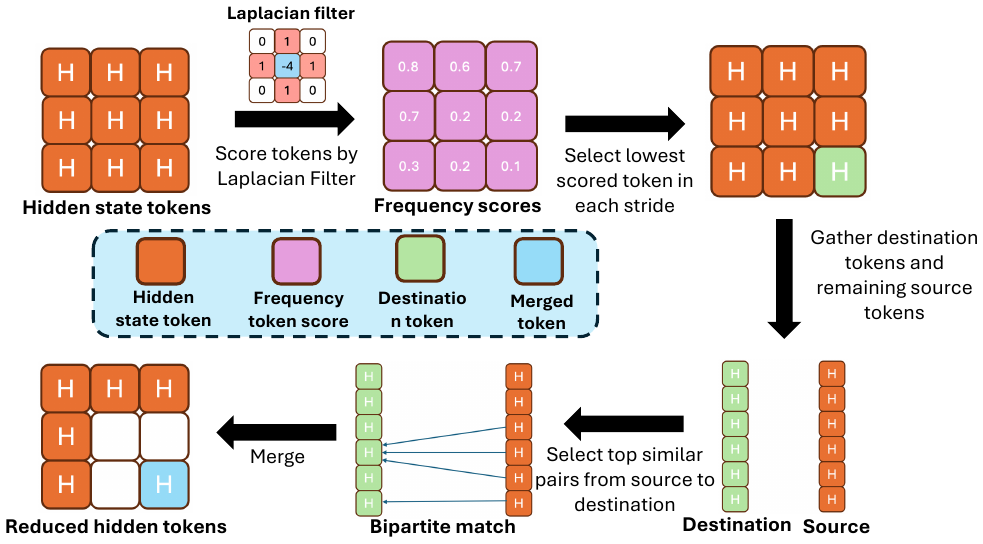}
  \vspace{-0.26in}
  \caption{Framework of our \textbf{BiGain\textsubscript{TM}} method. A Laplacian filter is applied to hidden-state tokens to compute local frequency scores. In each spatial stride, the lowest-scoring token is selected as a destination token, while the others form the source set. Destination and source tokens are gathered globally, and a bipartite matching selects top source-destination pairs.}
  \label{fig:framework}
\vspace{-0.1in}
\end{figure}

Diffusion models~\cite{ho2020denoising,song2020denoising,rombach2022high} have become the backbone of modern generative systems, yet their computational footprint during sampling has motivated a surge of training-free acceleration techniques such as token merging~\cite{bolya2023token} and token downsampling~\cite{smith2024todo}. Nearly all of these methods are evaluated primarily for generation/synthesis fidelity under reduced compute (e.g., keeping FID or perceptual quality stable while cutting FLOPs). This single-objective perspective overlooks an increasingly important use case: the same diffusion backbones are potentially and routinely repurposed for downstream recognition, either through linear probes on intermediate features, feature distillation into smaller classifiers~\cite{tang2023emergent,meng2024not}, or diffusion-based classification protocols~\cite{li2023your,clark2023text}. In practice, we observe that accelerations that {\em barely hurt} generation can dramatically weaken discriminative performance. This issue is critical as a single diffusion backbone can support both generation and classification through label-conditioned denoising likelihoods.

 Joint usage of generation and classification has been utilized across several application domains. In medical imaging, class-conditional diffusion models are used for both diagnostic prediction and counterfactual or uncertainty-aware reconstruction within one generative prior~\cite{favero2025conditional}. In safety-critical perception, robust diffusion classifiers employ the same denoiser for likelihood-based classification and image generation~\cite{chen2023robust}. In industrial visual inspection, diffusion backbones support defect identification and defect reconstruction using the same model~\cite{he2024diffusion,beizaee2025correcting}. In remote sensing, diffusion models are applied to cloud-removal or super-resolution synthesis while also improving land-cover and object classification~\cite{sousa2025data}.

\begin{figure}[t] 
  \centering
  \includegraphics[width=1.\linewidth]{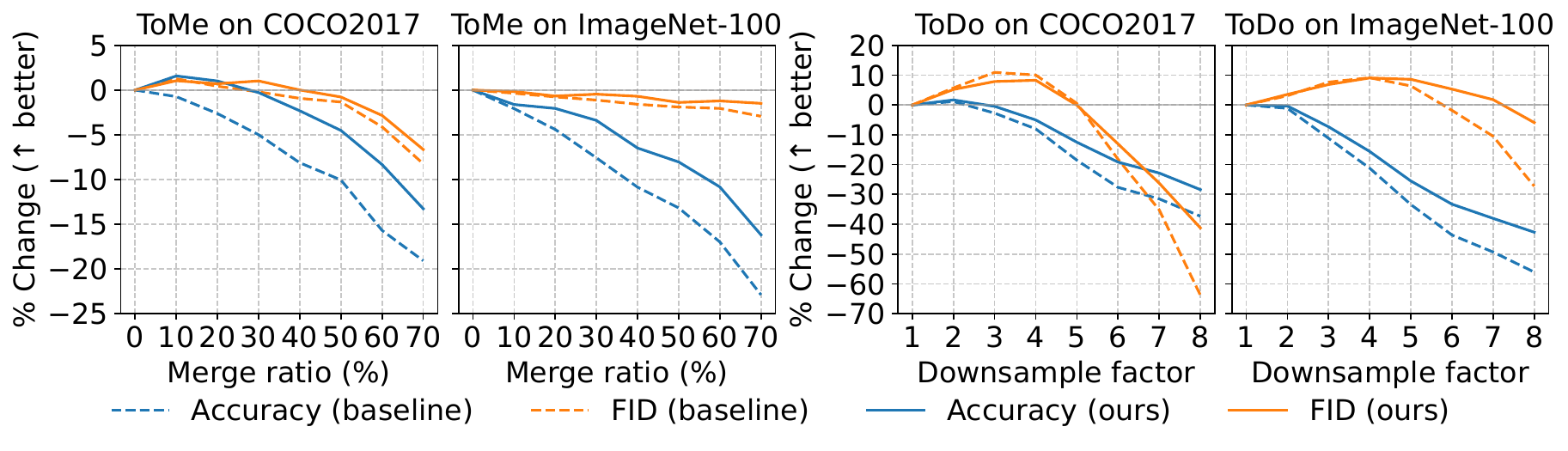}
  \vspace{-0.3in}
  \caption{
  Impact of token compression on diffusion models as our motivation on COCO2017 and ImageNet-100. {\bf Left:} ToMe~\cite{bolya2023token} (baseline) vs. Laplacian-Gated Merge (ours) as the merge ratio increases. {\bf Right:} ToDo~\cite{smith2024todo} (baseline) vs. Interpolate-Extrapolate KV-Downsampling (ours) as the downsample factor grows. Curves report percent change relative to the uncompressed model ($\uparrow$ better; for FID we plot $-\Delta$FID\%). {\bf Blue:} classification accuracy. {\bf Orange:} generation quality (FID). 
}
  \label{fig:motivation}
\vspace{-0.15in}
\end{figure}

Across these settings, generative and discriminative behaviors arise from one backbone, motivating that token compression methods need to preserve \emph{both} capabilities rather than optimizing solely for synthesis. Therefore, we argue that token compression should be rethought as a joint optimization problem that simultaneously safeguards generative fidelity and discriminative utility. Empirically, naive compression tends to remove precisely those structures that recognition benefits from (edge/texture cues, small objects, high-contrast boundaries), even when global appearance, and thus visual content remains complete. This creates a gap between what ``looks good'' and what ``classifies well''. To bridge this gap, we seek a compression principle that respects the complementary spectral needs of the two capabilities instead of privileging only synthesis. As shown in Fig.~\ref{fig:motivation}, baseline compression harms classification accuracy earlier and more sharply than synthesis, sometimes collapsing at extreme sparsity (e.g., COCO2017), whereas our approach consistently mitigates the accuracy drop while keeping generation competitive.

To reflect this, our key insight is frequency separation. Mapping signals of intermediate features into a frequency-aware representation disentangles high-frequency detail (edges, fine textures) from low-mid frequency content (shapes, layouts, semantics). This view yields a simple design rule: balanced spectral retention to preserve the high-frequency components that anchor recognition while maintaining the low-mid bands that support coherent generation. Guided by this principle, compression can prune redundancy without disproportionately harming either side.

In this work, we propose {\bf BiGain}, a training-free, plug-and-play framework composed of two operators. The first, {\em Laplacian-gated token merging}, computes local Laplacian magnitudes to guide merging: it encourages merges among spectrally smooth tokens and discourages merges of detail-carrying high-contrast tokens. This helps to retain edges and textured micro-structures that classifiers rely on, yet still collapses redundant flat regions to save compute. Crucially, the operator is architecture-agnostic and can be inserted at inference time without retraining. Second, {\em Interpolate-Extrapolate KV-downsampling} targets attention compute by downsampling keys/values with a controllable interpolation/extrapolation between nearest-neighbor and average pooling (IE-KVD), while leaving queries intact. Keeping queries at full resolution preserves the model’s ability to localize and attend precisely, whereas the KV shrinkage reduces memory and FLOPs smoothly, allowing a tunable speed-accuracy trade-off. Both methods are designed around the same principle of balanced spectral retention: preserving high-frequency detail while maintaining low and mid-frequency semantic content, enabling compression that respects both tasks.

Across DiT- and U-Net-based backbones and multiple datasets, BiGain consistently improves the speed-accuracy trade-off for diffusion-based classification while maintaining generation quality under comparable acceleration, often matching or slightly surpassing the synthesis fidelity of prior accelerations that do not consider recognition at all. Ablations confirm the necessity of frequency awareness: removing Laplacian gating disproportionately hurts classification, and downsampling KV in the frequency balance way is helpful for generation. These results suggest that respecting a balanced spectrum is a robust guiding principle for token compression.

Our contributions of this work are:
\begin{itemize}
    \item {\bf BiGain} reframes token compression for diffusion models as a bi-objective problem and offers a practical, training-free solution. 
    \item To our knowledge, it is the first framework to jointly study and advance both generation and classification under acceleration of diffusion models. 
    \item Beyond throughput and recognition gains, our study provides practical design guidance in a frequency-aware regime, merges where signals are smooth, downsamples $K,V$ while preserving $Q$ that informs future compression for deployable, dual-purpose generative models.
\end{itemize}

\section{Related Work}
\label{sec:related}

\noindent{\bf Acceleration of Diffusion Models.}
The iterative nature of diffusion models has spurred methods that reduce the \emph{number of steps} rather than alter the backbone. DDIM~\cite{song2020denoising} introduces non-Markovian sampling to take larger steps, and high-order solvers such as DPM-Solver~\cite{lu2022dpm} further shrink function evaluations while preserving fidelity. Progressive Distillation~\cite{salimansprogressive} compresses a teacher into a student that matches quality with fewer steps. These techniques largely treat the denoiser architecture as fixed and are thus orthogonal to our approach, which targets \emph{intra-step} compute via token compression. Meanwhile, pruning for diffusion~\cite{zhu2024dip,castells2024ld,fang2023structural} has also been explored. For example, Diff-Pruning~\cite{fang2023structural} uses a Taylor expansion over pruned timesteps, discarding non-contributory steps and aggregating informative gradients to rank important weights. DiP-GO~\cite{zhu2024dip} casts pruning as subnet search: it builds a SuperNet with backup connections over similar features and trains a plug-in pruner with tailored losses to identify redundant computation.

\noindent{\bf Token Reduction for Diffusion.}
Token reduction addresses the quadratic cost of attention by removing or merging redundant tokens. TokenLearner~\cite{ryoo2021tokenlearner} learns a small set of summary tokens, while training-free strategies like ToMe~\cite{bolya2023token} greedily merge similar tokens with minimal accuracy loss. Recent works adapt these ideas to diffusion backbones: ToMeSD~\cite{bolya2023tokenmergingfaststable} merges U-Net tokens at inference to accelerate Stable Diffusion, and complementary efforts explore structured pruning/sparsity for Diffusion Transformers~\cite{guoiccv2025,zhu2024dip}. Prior art primarily optimizes \emph{generation} speed-quality trade-offs and typically evaluates synthesis metrics; our method is also training-free and drop-in, but is explicitly designed to preserve generative fidelity \emph{and} discriminative utility through frequency-aware compression.

\noindent{\bf Diffusion as a Discriminative Learner, and the Open Gap.}
Beyond synthesis, diffusion models provide strong features for recognition~\cite{li2023your,clark2023text}. Diffusion-classifier frameworks use a pre-trained denoiser for per-class scoring or for feature extraction with a lightweight head, yielding competitive image classification~\cite{russakovsky2015imagenet,chen2023robust}. However, the interaction between \emph{token reduction} and \emph{discriminative performance} has been largely overlooked: accelerations that barely hurt synthesis can severely degrade classification. Our work sits at this intersection. We study how token compression affects both capabilities across U-Net/DiT backbones and introduce a frequency-aware, training-free framework that maintains generation quality while markedly improving diffusion-based classification.

\section{Methodology}
\label{sec:methods}

We first revisit token reduction for diffusion models from a \emph{bi-objective} viewpoint: preserve generative fidelity \emph{and} discriminative utility. After reviewing the denoising diffusion setup and the diffusion-classifier decision rule, we formalize shape-preserving token reduction and introduce two training-free, plug-in operators that are \emph{frequency-aware}: (i) {Laplacian-gated token merging} (L-GTM) and (ii) {Interpolate-Extrapolate KV-downsampling} (IE-KVD). Both operators avoid cross-timestep caching, which is incompatible with diffusion classification, and can be scheduled across timesteps/layers without retraining.

\subsection{Preliminaries}

A diffusion model~\cite{ho2020denoising,song2020denoising} specifies the forward process:
\begin{equation}
\label{eq:forward}
\begin{split}
q(\rvx_t \mid \vx_0)
= \mathcal{N}\!\bigl(\rvx_t ; \sqrt{\bar{\alpha}_t}\,\vx_0,\,(1-\bar{\alpha}_t)\,\mI\bigr),\\
\rvx_t
= \sqrt{\bar{\alpha}_t}\,\vx_0
  + \sqrt{1-\bar{\alpha}_t}\,\rvepsilon,
  \ \rvepsilon \sim \mathcal{N}(\vzero,\mI).
\end{split}
\end{equation}
where $\vx_0$ is the real clean data, $\rvx_t$ is its noisy version at step $t$, and $\rvepsilon$ is standard Gaussian noise. The scalar $\bar{\alpha}_t=\prod_{i=1}^t \alpha_i$ defines the cumulative noise schedule: a smaller $\bar{\alpha}_t$ means heavier corruption. Thus, each $\rvx_t$ is a linear combination of the original signal $\vx_0$ and the noise $\rvepsilon$.

The denoiser $\epsilon_{\vtheta}$ is trained in the noise-prediction parameterization:
\begin{equation}
\epsilon_{\vtheta}(\rvx_t,c,t) \approx \rvepsilon,
\quad
\mathcal{L}(\vtheta)=\E[\|\rvepsilon-\epsilon_{\vtheta}(\rvx_t,c,t)\|_2^2],
\end{equation}
where $c$ denotes an optional conditioning variable (e.g., class label or text prompt). 

The network $\epsilon_{\vtheta}$ learns to recover the exact Gaussian noise injected in the forward process. This training objective is equivalent to maximizing a variational lower bound (ELBO) on the data likelihood. It provides two core capabilities: (i) \emph{iterative generative sampling} by reversing the noising process, and (ii) \emph{per-class scoring for classification} by checking which conditioning $c$ yields the lowest prediction error.

\subsubsection{Diffusion Classifier} \label{D_classifier}
\textbf{Decision rule.} Given \(\vx\) and class set $\mathcal{C}$, draw a \emph{shared} Monte Carlo set \(\mathcal{S}_{\text{MC}}\!\!=\!\!\{(t_s,\rvepsilon_s)\}_{s=1}^S\) for all classes. Define:
\begin{equation}
\begin{split}
\rvx_{t_s}
= \sqrt{\bar{\alpha}_{t_s}}\,\vx
  + \sqrt{1-\bar{\alpha}_{t_s}}\,\rvepsilon_s,\\
\ell(\vx,c;t_s,\rvepsilon_s)
= \bigl\|\rvepsilon_s
   - \epsilon_{\vtheta}(\rvx_{t_s},c,t_s)\bigr\|_2^2 .
\end{split}
\end{equation}
Here $\bar{\alpha}_{t_s}$ and $\rvepsilon_s$ are as defined in the diffusion setup above, and $\epsilon_{\vtheta}$ is the same denoiser evaluated under class conditioning $c$. Thus $\ell(\vx,c;t_s,\rvepsilon_s)$ quantifies how well conditioning on $c$ explains the corruption realized at $(t_s,\rvepsilon_s)$. The class score and prediction are:
{\small
\begin{equation}
\widehat{L}(\vx,c)=\tfrac{1}{S}\!\sum_{s=1}^S \ell(\vx,c;t_s,\rvepsilon_s),
\hat{y}(\vx)=\argmin_{c\in\mathcal{C}}\,\widehat{L}(\vx,c).
\end{equation}
}
Sharing \((t_s,\rvepsilon_s)\) across classes yields a paired-difference estimate of the ELBO for \(\log p_{\vtheta}(\vx\!\mid\! c)\) without changing the decision rule.

\noindent{\bf Adaptive evaluation (staged pruning).}
For large $|\mathcal{C}|$, uniform evaluation is costly. We therefore allocate computation in $N_{\mathrm{stages}}$ rounds with cumulative budgets $(T_1,\dots,T_{N_{\mathrm{stages}}})$ and keep-counts $(K_1,\dots,K_{N_{\mathrm{stages}}})$ (also see Appendix): at stage $i$, each surviving class accrues evaluations up to $T_i$, then only the $K_i$ lowest-score classes are retained for the next stage. This staged pruning discards unlikely classes early and concentrates samples on plausible ones, reducing wall-clock compute while leaving the final decision $\argmin_{c}\widehat{L}(\vx,c)$ unchanged.

\subsubsection{Attention and Shape-Preserving Token Reduction}

Let the denoiser operate on $N$ latent tokens \(\mX\in\R^{N\times d}\) (rows \(\vx_i\)). A standard self-attention block forms:
\begin{equation}
\begin{split}
\mQ = \mX \mW_Q,\quad \mK = \mX \mW_K,\quad \mV = \mX \mW_V,\\[4pt]
\operatorname{Attn}(\mX)
= \softmax\!\left(\frac{\mQ \mK^\top}{\sqrt{d_k}}\right)\mV .
\end{split}
\end{equation}
To accelerate while keeping the output length $N$, we use a shape-preserving reduction $\mM \in \R^{N'\times N}$ with $N'\!<\!N$, and, if queries are reduced, an unmerge operator $\mU \in \R^{N\times N'}$:
\begin{equation}
\mX \xrightarrow{\tilde{\mX}=\mM\mX} \mZ=F(\tilde{\mX}) \xrightarrow{\bar{\mX}=\mU\mZ} \bar{\mX}\in\R^{N\times d}.
\end{equation}
We consider two concrete, training-free instances below.

\subsection{BiGain: Frequency-Aware Token Compression}
\label{sec:bigain}
Our central design rule is \textbf{balanced spectral retention}: preserve high-frequency detail (edges/textures) and low/mid-frequency content (global semantics). We instantiate this via two complementary operators.

\subsubsection{Laplacian-Gated Token Merging (L-GTM)}
\label{sec:lgtm}
\textbf{Goal.}  In this part, we aim to merge spectrally smooth tokens while discouraging merges of high-contrast tokens.

\noindent\textbf{Spectral proxy.} We reshape \(\mX\in\R^{N\times d}\) to \(\mX\in\R^{H\times W\times C}\) ($C=d$) and compute a per-location frequency magnitude via a spatial Laplacian:
{\small
\begin{equation}
\mF=\operatorname{Reduce}_c\!\bigl(|\mX * \mL|\bigr),\quad
\mL=\begin{bmatrix}0&1&0\\[2pt]1&-4&1\\[2pt]0&1&0\end{bmatrix},\quad
\mF\in\R^{H\times W}.
\end{equation}
}
Here $\operatorname{Reduce}_c$ is channel-wise aggregation (e.g., mean or $\ell_2$). $\mL$ denotes the {\em Laplacian kernel}, a finite approximation of the second-order derivatives of features in the spatial dimensions (height and width). It is used to measure the degree of difference with respect to the local neighborhood.

\noindent\textbf{Gated merging.}
Let $s_{ij}=\mF_{ij}$, in each grid, we take the tokens with the lowest $s_{ij}$ values as the destination set $\mathcal{A}$ (low-frequency anchors), and all remaining tokens as the source set $\mathcal{B}$. We then merge the top $r\%$ most similar source-destination pairs by equal-weight averaging. The resulting anchors form the reduced sequence $\tilde{\mX}$, which defines the merge operator $\mM$; if needed, $\mU$ restores shape by broadcasting averaged values back to removed indices. This encourages compression among spectrally smooth tokens while leaving high-frequency tokens largely intact.

\noindent\textbf{Compute.} When \(\mM\) reduces \(\mQ,\mK,\mV\) to $N$' tokens, attention costs shrink from $\mathcal{O}(N^2d)$ to $\mathcal{O}(N'^2 d)$. L-GTM is architecture-agnostic and training-free, we never touch class tokens in DiT nor time/text conditioning tokens in U-Net cross-attention.

\noindent{\bf Blockwise ABM (Adaptive Block Merging)}: a fast variant. For additional efficiency, we introduce a tiled variant that pools an $s\times s$ block $t$ only if $\phi(t)=\max_{(i,j)\in t}\mF_{ij}<\tau$ (with $\tau$ as a quantile of $\mF$). Pooled tokens are averaged, others are kept verbatim. ABM is a drop-in replacement for L-GTM in high-resolution stages.

\subsubsection{Interpolate--Extrapolate KV-Downsampling (IE-KVD)}
\label{sec:ikvd}

\noindent\textbf{Goal.} In this component, we aim to reduce attention cost by downsampling keys/values while keeping queries intact to preserve localization and alignment.

\noindent\textbf{Operator.} Given a stride $s$ and reduced grid $\tilde{H}\times\tilde{W} (\tilde{N}=\tilde{H}\tilde{W}\ll N)$, we define a per-site interpolator/extrapolator between nearest and average pooling:
\begin{equation}
\label{eq:kv_operator}
\begin{split}
\mathcal{D}_{\alpha,s}(\mZ)[i]
= \alpha\,\mZ[\text{nearest}(i)]
  + (1{-}\alpha)\,\tfrac{1}{|\mathcal{N}_s(i)|}
    \sum_{j \in \mathcal{N}_s(i)} \mZ[j],\\[3pt]
\end{split}
\end{equation}
We set \(\tilde{\mK}=\mathcal{D}_{\alpha,s}(\mK)\) and \(\tilde{\mV}=\mathcal{D}_{\alpha,s}(\mV)\), while \(\mQ\) remains full-resolution. The attention then costs $\mathcal{O}(N\tilde{N}d)$ and preserves output length $N$.

Preserving \(\mQ\) maintains fine-grained receptive fields for every output token, which stabilizes synthesis, and critically retains discriminative cues (edge/texture) in diffusion classification, where per-token attention precision matters for the MC scoring rule.

\subsection{Compatibility with Diffusion Classification}
Our operators are \emph{timestep-local}, deterministic given \(\mX\), and do not rely on cross-timestep caches. They therefore integrate seamlessly with the diffusion-classifier decision rule in Sec.~\ref{D_classifier}: all classes receive identical \((t_s,\rvepsilon_s)\) and identical compression schedules, so the paired-difference estimator remains valid. In practice, we reduce per-class FLOPs \emph{and} improve accuracy relative to baselines that focus solely on synthesis quality.

\vspace{-0.05in}
\section{Experiments}

\subsection{Experimental Setup}
\noindent \textbf{Models.}
We test our {\bf BiGain} on two representative diffusion models: {\bf Stable Diffusion v2.0}~\cite{rombach2022high} (UNet-based latent diffusion with text conditioning) and {\bf DiT-XL/2} \cite{Peebles2022DiT} (Transformer backbone), using official pretrained weights. Diffusion classifiers require a noise predictor $\hat\epsilon_\theta(x_t,t)$. 

\noindent{\bf Datasets and Metrics.} For the classification task,  we evaluate on four representative datasets: {\texttt{ImageNet-1K}~\cite{russakovsky2015imagenet}, \texttt{ImageNet-100}~\cite{tian2020contrastive}, \texttt{Oxford-IIIT Pets}~\cite{parkhi2012cats}, and \texttt{COCO2017}~\cite{lin2014microsoft}}. Following~\cite{li2023your}, we evaluate on a 2,000-image validation subset for ImageNet-1K (linear cost in $|\mathcal C|$), full validation splits are used elsewhere. We report Top-1 accuracy for single-label datasets and Top-1 precision plus mAP (macro) for multi-label COCO. For generation we evaluate on COCO2017 captions, ImageNet-100, and ImageNet-1K, reporting FID metric. DiT-XL/2 is evaluated only on ImageNet datasets (class-index conditioning, no free-form prompts), while Stable Diffusion v2.0 is evaluated on all datasets using text class prompts. Efficiency is reported as sparsity and FLOPs (both total and attention FLOPs). More details are provided in our appendix.

\begin{table*}[t]
\centering
\caption{Classification accuracy (Acc@1) on Pets dataset under similar FLOPs reduction.}
\label{tab:pets-10x}
\vspace{-0.1in}
\small
\resizebox{.8\textwidth}{!}{
\begin{tabular}{lcccc}
\toprule
  Method &Acceleration Type&  FLOPs Reduction $\uparrow$ & Acc@1 $\uparrow$ (\%) & $\Delta$ vs. Baseline $\downarrow$ \\
\midrule
Baseline (No Accel.) &None& -- & 81.03 & -- \\
\midrule
ToMe~\cite{bolya2023token} & Token Merging/Pruning  & $10 $\% & 72.96 & $\downarrow$ 8.07\\
DiP-GO~\cite{zhu2024dip}  & Model Pruning&$10$\% & 76.53 & $\downarrow$ 4.50 \\
SiTo~\cite{zhang2025training}& Token Merging/Pruning & $7 $\% & 68.84 & $\downarrow$ 12.19 \\
MosaicDiff~\cite{guoiccv2025}  &Model Pruning          & $10$\% & 77.38 & $\downarrow$ 3.65 \\
\textbf{BiGain$_\texttt{TM}$ (Ours)}&Token Merging/Pruning  & \cellcolor{lightgray!50}$10\%$ & \cellcolor{lightgray!50}\textbf{78.38} & \cellcolor{lightgray!50}\textbf{$\downarrow$ 2.65} \\ \midrule
ToDo \cite{smith2024todo} &Token Downsampling &  $14.2 $\% &  79.15  & $\downarrow$ 1.88    \\
\textbf{BiGain$_\texttt{TD}$ (Ours)} & Token Downsampling & \cellcolor{lightgray!50}$14.2\%$ & \cellcolor{lightgray!50}\textbf{79.90} & \cellcolor{lightgray!50}\textbf{$\downarrow$ 1.13} \\
\bottomrule
\end{tabular}
}
\end{table*}

\begin{table*}[h]
\centering
\caption{SD-2.0 \textbf{Token Downsampling}: Classification (Acc@1 on Pets, ImageNet-100; Acc@1 and mAP on COCO-2017) and generation fidelity (FID $\downarrow$) vs. downsampling factor. For classification, we fix the interextrapolation factor at 0.9 across all timesteps to ensure stability. For generation, we linearly vary the factor from 0.8 (early steps) to 1.2 (later steps), shifting emphasis from low- to high-frequency information. {\color{gray}{Gray}} color indicates the same generation results as the above group.}
\label{tab:sd20-down_classification}
\vspace{-0.07in}
\small
\setlength{\tabcolsep}{5pt}
\resizebox{.9\textwidth}{!}{
\begin{tabular}{l c ccccccc c ccc}
\toprule
 \multirow{2}{*}{Method} & \multirow{2}{*}{No Accel.} &  \multicolumn{7}{c}{\textbf{Classification $\uparrow$ (TD$\times$)}} &\multirow{2}{*}{No Accel.}   &\multicolumn{3}{c}{\bf Generation $\downarrow$ (TD$\times$)}  \\
     &      & 2$\times$  & 3$\times$ & 4$\times$ & 5$\times$ & 6$\times$ & 7$\times$ &8$\times$ & & 2$\times$  & 3$\times$ & 4$\times$   \\
\midrule
\multicolumn{13}{c}{\textbf{Pets}}  \\
Avg-pooling (baseline)     & &77.02 & 73.45 & 71.26 & 69.00 & 67.56 & 66.66 & 65.13    &\multirow{3}{*}{35.01}&38.50&39.42&39.74\\
   ToDo \cite{smith2024todo}                     & 81.03 & 81.30 & 79.15 & 77.46 & 72.74 & 66.74 & 62.87 & 56.16&  &33.52&32.38&31.48\\
 \textbf{BiGain$_\texttt{TD}$}(Ours)    & \textbf{} & \textbf{81.52} & \textbf{79.91} & \textbf{78.03} & \textbf{74.92} & \textbf{70.86} & \textbf{69.33} & \textbf{66.03} 
  & &\textbf{32.19}&\textbf{30.44}&\textbf{29.21}\\
    \cdashlinelr{1-13}
     {\color{Peach}{$\Delta\uparrow$}}   &           &   {\color{Peach}{$\uparrow$0.22}}             &     {\color{Peach}{$\uparrow$0.76}}            &      {\color{Peach}{$\uparrow$0.57}}          &       {\color{Peach}{$\uparrow$2.18}}          &     {\color{Peach}{$\uparrow$4.12}}            &     {\color{Peach}{$\uparrow$6.46}}            &       {\color{Peach}{$\uparrow$9.87}}           & &
  {\color{Peach}{$\downarrow$1.33}} 
  &{\color{Peach}{$\downarrow$1.94}} &
  {\color{Peach}{$\downarrow$2.27}} \\
\midrule
\multicolumn{2}{c}{} &\multicolumn{7}{c}{\textbf{\hspace{-3.4em}ImageNet-100}} & \multicolumn{1}{c}{} &\multicolumn{3}{c}{\textbf{ImageNet-1K}}  \\
  Avg-pooling (baseline)       & \multirow{3}{*}{73.12} & 58.50 & 49.52 & 45.54 & 40.96 & 38.74 & 38.12 & 37.40 & \multirow{3}{*}{17.64} & 19.31 & 23.08  & 26.59   \\
   ToDo \cite{smith2024todo}                     &  & 72.30 & 64.96 & 57.62 & 48.70 & 41.22 & 37.04 & 32.12  &  & 16.86  & 15.93 & 15.63  \\
   \textbf{BiGain$_\texttt{TD}$}(Ours)    & & \textbf{72.88} & \textbf{67.78} & \textbf{61.72} & \textbf{54.48} & \textbf{48.78} & \textbf{45.30} & \textbf{41.90} &  & \textbf{16.46}  & \textbf{15.46} & \textbf{15.46}      \\
    \cdashlinelr{1-13}
     {\color{Peach}{$\Delta\uparrow$}}   &           &   {\color{Peach}{$\uparrow$0.58}}             &     {\color{Peach}{$\uparrow$2.82}}            &      {\color{Peach}{$\uparrow$4.10}}          &       {\color{Peach}{$\uparrow$5.78}}          &     {\color{Peach}{$\uparrow$7.56}}            &     {\color{Peach}{$\uparrow$8.26}}            &       {\color{Peach}{$\uparrow$9.78}}           & &
  {\color{Peach}{$\downarrow$0.40}} 
  &{\color{Peach}{$\downarrow$0.47}} &
  {\color{Peach}{$\downarrow$0.17}} \\
\midrule
\multicolumn{13}{c}{\textbf{COCO-2017}}  \\
\emph{Acc@1} Avg-pooling (baseline)            &  & 62.98 & 55.94 & 52.46 & 48.74 & 46.88 & 47.38 & 46.74 & \multirow{3}{*}{26.79} & 30.52  & 35.92 & 41.23  \\
   \emph{Acc@1} ToDo \cite{smith2024todo}                      & 70.84 & 71.66 & 68.90 & 65.16 & 57.70 & 51.26 & 48.52 & 44.40 &  & 25.26 & 23.86 & 24.10  \\
   \emph{Acc@1} \textbf{BiGain$_\texttt{TD}$}(Ours)    & \textbf{} & \textbf{72.04} & \textbf{70.52} & \textbf{67.28} & \textbf{61.98} & \textbf{57.26} & \textbf{54.66} & \textbf{50.72} & & \textbf{24.29} & \textbf{23.17} & \textbf{24.05}   \\
    \cdashlinelr{1-13}
     {\color{Peach}{$\Delta\uparrow$}}   &           &   {\color{Peach}{$\uparrow$0.38}}             &     {\color{Peach}{$\uparrow$1.62}}            &      {\color{Peach}{$\uparrow$2.12}}          &       {\color{Peach}{$\uparrow$4.28}}          &     {\color{Peach}{$\uparrow$6.00}}            &     {\color{Peach}{$\uparrow$6.14}}            &       {\color{Peach}{$\uparrow$6.32}}           & & {\color{Peach}{$\downarrow$0.97}}   &{\color{Peach}{$\downarrow$0.69}} & {\color{Peach}{$\downarrow$0.05}} \\
  \cmidrule(lr){1-13}
   \emph{mAP} Avg-pooling (baseline)                     &  & 44.25 & 40.96 & 38.98 & 36.89 & 35.77 & 35.79 & 35.38 &\cellcolor{lightgray!50}\strut &\cellcolor{lightgray!50}30.52&\cellcolor{lightgray!50}35.92&\cellcolor{lightgray!50}41.23 \\
   \emph{mAP}  ToDo \cite{smith2024todo}                       & 46.01 & 46.59 & 45.56 & 44.07 & 40.31 & 36.95 & 35.50 & 33.34 &\cellcolor{lightgray!50}26.79 &\cellcolor{lightgray!50}25.26&\cellcolor{lightgray!50}23.86&\cellcolor{lightgray!50}24.10 \\
   \emph{mAP} \textbf{BiGain$_\texttt{TD}$}(Ours)      & \textbf{} & \textbf{46.97} & \textbf{46.28} & \textbf{44.81} & \textbf{42.54} & \textbf{40.28} & \textbf{38.82} & \textbf{36.93} & \cellcolor{lightgray!50}\strut &\cellcolor{lightgray!50}\textbf{24.29}&\cellcolor{lightgray!50}\textbf{23.17}&\cellcolor{lightgray!50}\textbf{24.05} \\
    \cdashlinelr{1-13}

     {\color{Peach}{$\Delta\uparrow$}}   &           &   {\color{Peach}{$\uparrow$0.38}}             &     {\color{Peach}{$\uparrow$0.72}}            &      {\color{Peach}{$\uparrow$0.74}}          &       {\color{Peach}{$\uparrow$2.23}}          &     {\color{Peach}{$\uparrow$3.33}}            &     {\color{Peach}{$\uparrow$3.32}}            &       {\color{Peach}{$\uparrow$3.59}}         & \cellcolor{lightgray!50}\strut  & \cellcolor{lightgray!50}{\color{Peach}{$\downarrow$0.97}}   &\cellcolor{lightgray!50}{\color{Peach}{$\downarrow$0.69}} &\cellcolor{lightgray!50} {\color{Peach}{$\downarrow$0.05}} \\
\bottomrule
\end{tabular}
}

\vspace{0.1in}
\centering
\small
\caption{DiT-XL/2 \textbf{Token Downsampling}: Classification (Acc@1) and generation fidelity (FID $\downarrow$) vs. downsampling factor. For both classification and generation, we fix the interpolate-extrapolate factor at 0.1 across all timesteps. TD Factor: Token Downsampling factor.}
\label{tab:dit-down_classification}
\vspace{-0.1in}
\setlength{\tabcolsep}{5pt}
\resizebox{0.9\textwidth}{!}{
\begin{tabular}{lccccccccccc}
\toprule
 \multirow{2}{*}{Method} & \multirow{2}{*}{No Accel.} &  \multicolumn{4}{c}{\textbf{Classification $\uparrow$ (TD$\times$)}} &\multirow{2}{*}{No Accel.}   &\multicolumn{4}{c}{\bf Generation $\downarrow$ (TD$\times$)}  \\
  &           & 2$\times$  & 3$\times$ & 4$\times$ & 5$\times$ & & 2$\times$  & 3$\times$ & 4$\times$ & 5$\times$  \\
\midrule
\multicolumn{10}{c}{\textbf{ImageNet-100}}  \\
Avg-pooling (baseline)       & \multirow{3}{*}{84.82} & 78.34 & 61.04 & 48.40 & 33.26 & \multirow{3}{*}{47.53}&\textbf{40.13}&33.57&30.25&41.61 \\
   ToDo~\cite{smith2024todo}                   &  & 69.34 & 8.46 & 4.74 & 3.32 & &40.48&190.18&206.52&215.04 \\
   \textbf{BiGain$_\texttt{TD}$ (Ours)}   & \textbf{} & \textbf{78.42} & \textbf{61.58} & \textbf{48.72} & \textbf{34.00}   &&\textbf{40.13}&\textbf{32.95}&\textbf{29.87}&\textbf{40.55}\\
\cdashlinelr{1-11}
  {\color{Peach}{$\Delta\uparrow$}}   &           
&   {\color{Peach}{$\uparrow$9.08}}  &   {\color{Peach}{$\uparrow$53.12}}  
&   {\color{Peach}{$\uparrow$43.98}}  &   {\color{Peach}{$\uparrow$30.68}}  
&   &   {\color{Peach}{$\downarrow$0.35}}  &   {\color{Peach}{$\downarrow$157.23}}  
&   {\color{Peach}{$\downarrow$176.65}}  &   {\color{Peach}{$\downarrow$174.49}} \\
\bottomrule
\end{tabular}
}
\vspace{-0.1in}
\end{table*}

\begin{table*}[h]
\centering
\caption{SD-2.0 \textbf{Token Merging}: Classification (Acc@1 on Pets, ImageNet-100/1K; Acc@1 and mAP on COCO-2017) and generation fidelity (FID $\downarrow$) vs. Token Merging Ratio.}
\label{tab:sd20-merge_classification}
\vspace{-0.1in}
\small
\resizebox{0.97\textwidth}{!}{
\begin{tabular}{lcccccccccccccccc}
\toprule
 \multirow{2}{*}{Method} & \multirow{2}{*}{No Accel.} &  \multicolumn{7}{c}{\textbf{Classification  $\uparrow$ (Token Merging Ratio)}} &\multirow{2}{*}{No Accel.}   &\multicolumn{7}{c}{\bf Generation $\downarrow$ (Token Merging Ratio)}  \\
     &      & 10\% & 20\% & 30\% & 40\% & 50\% & 60\% & 70\% & & 10\% & 20\% & 30\% & 40\% & 50\% & 60\% & 70\%  \\
\midrule
\multicolumn{17}{c}{\textbf{Pets}}  \\
ToMe & \multirow{3}{*}{81.03} & 80.10 & 79.88 & 78.44 & 76.42 & 72.96 & 69.93 & 65.76  &  & 35.05 & 35.30& 35.71&36.26 & 37.00& 37.63 & 38.35 \\
   \textbf{BiGain$_\texttt{TM}$ (Ours)} & \textbf{} & \textbf{81.16} & \textbf{81.16} & \textbf{80.40} & \textbf{80.07} & \textbf{78.38} & \textbf{76.04} & \textbf{74.63}  & 35.01& \textbf{35.00} & \textbf{35.12}& \textbf{35.01} &\textbf{35.99} &\textbf{36.52} & \textbf{36.99}&\textbf{37.73}\\ 
  \cdashlinelr{1-13}
    {\color{Peach}{$\Delta\uparrow$}}  &    & {\color{Peach}{$\uparrow$1.06}}   & {\color{Peach}{$\uparrow$1.28}} &{\color{Peach}{$\uparrow$1.96}} & {\color{Peach}{$\uparrow$3.65}} & {\color{Peach}{$\uparrow$5.42}}  & {\color{Peach}{$\uparrow$6.11}}& {\color{Peach}{$\uparrow$8.87}}&    & 
    
    {\color{Peach}{$\downarrow$0.05}}   & {\color{Peach}{$\downarrow$0.18}} &{\color{Peach}{$\downarrow$0.70}} & {\color{Peach}{$\downarrow$0.27}} & {\color{Peach}{$\downarrow$0.48}}  & {\color{Peach}{$\downarrow$0.64}}& {\color{Peach}{$\downarrow$0.62}}  \\
\midrule
\multicolumn{17}{c}{\textbf{ImageNet-100}}  \\
  ToMe      & \multirow{3}{*}{73.12} & 71.60 & 69.90 & 67.58 & 65.18 & 63.48 & 60.70 & 56.38 &  
  &41.51&41.68&41.82&42.02&42.15&42.21&42.58 \\
   \textbf{BiGain$_\texttt{TM}$ (Ours)} & & \textbf{71.94} & \textbf{71.62} & \textbf{70.64} & \textbf{68.38} & \textbf{67.24} & \textbf{65.20} & \textbf{61.28}  & 41.37
  &\textbf{41.43}&\textbf{41.64}&\textbf{41.55}&\textbf{41.65}&\textbf{41.93}&\textbf{41.86}&\textbf{41.98} \\ \cdashlinelr{1-13}
   {\color{Peach}{$\Delta\uparrow$}}   &           &   {\color{Peach}{$\uparrow$0.34}}             &     {\color{Peach}{$\uparrow$1.72}}            &      {\color{Peach}{$\uparrow$3.06}}          &       {\color{Peach}{$\uparrow$3.20}}          &     {\color{Peach}{$\uparrow$3.76}}            &     {\color{Peach}{$\uparrow$4.50}}            &       {\color{Peach}{$\uparrow$4.90}}           &    & 
    
    {\color{Peach}{$\downarrow$0.08}}   & {\color{Peach}{$\downarrow$0.04}} &{\color{Peach}{$\downarrow$0.27}} & {\color{Peach}{$\downarrow$0.37}} & {\color{Peach}{$\downarrow$0.22}}  & {\color{Peach}{$\downarrow$0.35}}& {\color{Peach}{$\downarrow$0.60}}
   \\
\midrule
\multicolumn{17}{c}{\textbf{ImageNet-1K}}  \\
ToMe     & \multirow{3}{*}{57.05} & 55.50 & 54.25 & 52.35 & 50.65 &  47.55  & 43.55  & 37.35 &\multirow{3}{*}{17.64} & 17.57  & 17.66 &  17.74 & 17.74  &  17.83 & 17.97 &  18.42  \\
   \textbf{BiGain$_\texttt{TM}$ (Ours)} & & \textbf{57.25} & \textbf{56.50} & \textbf{55.80 } & \textbf{54.80} & \textbf{52.50} & \textbf{49.10} & \textbf{44.50 } & & \textbf{17.54}
 & \textbf{17.48} & \textbf{17.52} & \textbf{17.53} & \textbf{17.58} & \textbf{17.69} & \textbf{18.08}  \\  \cdashlinelr{1-13}
{\color{Peach}{$\Delta\uparrow$}} &           &         {\color{Peach}{$\uparrow$1.75}}       &       {\color{Peach}{$\uparrow$2.25}}         &          {\color{Peach}{$\uparrow$3.45}}     &          {\color{Peach}{$\uparrow$4.15}}      &   {\color{Peach}{$\uparrow$4.95}}             &         {\color{Peach}{$\uparrow$5.55}}       &      {\color{Peach}{$\uparrow$7.15}}           &   & 
    
    {\color{Peach}{$\downarrow$0.03}}   & {\color{Peach}{$\downarrow$0.18}} &{\color{Peach}{$\downarrow$0.22}} & {\color{Peach}{$\downarrow$0.21}} & {\color{Peach}{$\downarrow$0.25}}  & {\color{Peach}{$\downarrow$0.28}}& {\color{Peach}{$\downarrow$0.34}}
   \\
\midrule
\multicolumn{17}{c}{\textbf{COCO-2017}}  \\
\emph{Acc@1} $|$ ToMe    & \multirow{3}{*}{70.84} & 70.32 & 68.98 & 67.3 & 65.08 & 63.72 & 59.72 & 57.32 & \multirow{3}{*}{26.79}  & \textbf{26.45} &26.68 &26.85 &27.04 &27.15 &27.89 &29.00 \\
   \emph{Acc@1} $|$ \textbf{BiGain$_\texttt{TM}$ } & \textbf{} & \textbf{71.96} & \textbf{71.56} & \textbf{70.64} & \textbf{69.20} & \textbf{67.64} & \textbf{64.94} & \textbf{61.44}  &  &  26.51 &\textbf{26.60} &\textbf{26.52} &\textbf{26.79} &\textbf{27.00} &\textbf{27.55} &\textbf{28.57} \\ 
  \cdashlinelr{2-17}
   {\color{Peach}{$\Delta\uparrow$}} &           &  {\color{Peach}{$\uparrow$1.64}}          &     {\color{Peach}{$\uparrow$2.58}}           &          {\color{Peach}{$\uparrow$3.34}}    &          {\color{Peach}{$\uparrow$4.12}}      &   {\color{Peach}{$\uparrow$3.92}}             &        {\color{Peach}{$\uparrow$5.22}}        &     {\color{Peach}{$\uparrow$4.12}}            &   & 
    {\color{Peach}{$\uparrow$0.06}}   & {\color{Peach}{$\downarrow$0.08}} &{\color{Peach}{$\downarrow$0.33}} & {\color{Peach}{$\downarrow$0.25}} & {\color{Peach}{$\downarrow$0.15}}  & {\color{Peach}{$\downarrow$0.34}}& {\color{Peach}{$\downarrow$0.43}}\\
  \cmidrule(lr){2-17}
   \emph{mAP} $|$ ToMe   & \multirow{3}{*}{46.01} & 46.04 & 45.35 & 44.50 & 43.43 & 42.82 & 41.01 & 40.07 &\cellcolor{lightgray!50} &\cellcolor{lightgray!50}\textbf{26.45} &\cellcolor{lightgray!50}26.68 &\cellcolor{lightgray!50}26.85 &\cellcolor{lightgray!50}27.04 &\cellcolor{lightgray!50}27.15 &\cellcolor{lightgray!50}27.89 &\cellcolor{lightgray!50}29.00\\
   \emph{mAP} $|$ \textbf{BiGain$_\texttt{TM}$ }   & \textbf{} & \textbf{46.38} & \textbf{46.21} & \textbf{46.05} & \textbf{45.50} & \textbf{44.94} & \textbf{43.98} & \textbf{42.44} &\cellcolor{lightgray!50}26.79 &\cellcolor{lightgray!50}26.51 &\cellcolor{lightgray!50}\textbf{26.60} &\cellcolor{lightgray!50}\textbf{26.52} &\cellcolor{lightgray!50}\textbf{26.79} &\cellcolor{lightgray!50}\textbf{27.00} &\cellcolor{lightgray!50}\textbf{27.55} &\cellcolor{lightgray!50}\textbf{28.57} \\ 
  \cdashlinelr{2-17}
   {\color{Peach}{$\Delta\uparrow$}}&           &   {\color{Peach}{$\uparrow$0.34}}           &          {\color{Peach}{$\uparrow$0.86}}    &          {\color{Peach}{$\uparrow$1.55}}      &   {\color{Peach}{$\uparrow$2.07}}             &        {\color{Peach}{$\uparrow$2.12}}        &     {\color{Peach}{$\uparrow$2.97}} & {\color{Peach}{$\uparrow$2.37}}          & \cellcolor{lightgray!50}  &  \cellcolor{lightgray!50}  {\color{Peach}{$\uparrow$0.06}}   &\cellcolor{lightgray!50} {\color{Peach}{$\downarrow$0.08}} &\cellcolor{lightgray!50}{\color{Peach}{$\downarrow$0.33}} & \cellcolor{lightgray!50}{\color{Peach}{$\downarrow$0.25}} & \cellcolor{lightgray!50}{\color{Peach}{$\downarrow$0.15}}  & \cellcolor{lightgray!50}{\color{Peach}{$\downarrow$0.34}}& \cellcolor{lightgray!50}{\color{Peach}{$\downarrow$0.43}}  \\
\bottomrule
\end{tabular}
}

\vspace{0.15in}
\centering
\caption{DiT-XL/2 \textbf{Token Merging}: Classification (Acc@1) and generation fidelity (FID $\downarrow$) vs. Token Merging Ratio.}
\label{tab:dit_merge_cls_gen}
\vspace{-0.07in}
\setlength{\tabcolsep}{6pt}
\resizebox{1.0\textwidth}{!}{
\begin{tabular}{lcccccccccccccccc}
\toprule
 \multirow{2}{*}{Method} & \multirow{2}{*}{No Accel.} &  \multicolumn{7}{c}{\textbf{Classification $\uparrow$ (Token Merging Ratio)}} &\multirow{2}{*}{No Accel.}   &\multicolumn{7}{c}{\bf Generation $\downarrow$ (Token Merging Ratio)}  \\
   &      & 10\% & 20\% & 30\% & 40\% & 50\% & 60\% & 70\% & & 10\% & 20\% & 30\% & 40\% & 50\% & 60\% & 70\%  \\
\midrule
\multicolumn{17}{c}{\textbf{ImageNet-100}}  \\
  ToMe            & \multirow{3}{*}{84.82} & 80.86 & 78.02 & 75.3 & 71.38 & 68.24 & 62.06 & 53.88 & &41.51 & 41.68 & 41.83 & 42.02 & 42.15 & 42.21 & 42.58\\
    \textbf{BiGain$_\texttt{TM}$} & \textbf{} & \textbf{83.56} & \textbf{82.2} & \textbf{79.92} & \textbf{77.38} & \textbf{73.68} & \textbf{68.34} & \textbf{61.76} & 47.53
    &\textbf{41.43} & \textbf{41.61} & \textbf{41.56} & \textbf{41.65} & \textbf{41.92} & \textbf{41.77} & \textbf{41.89} \\
\cdashlinelr{1-13}
  {\color{Peach}{$\Delta\uparrow$}}   &           
&   {\color{Peach}{$\uparrow$2.70}}  &   {\color{Peach}{$\uparrow$4.18}}  
&   {\color{Peach}{$\uparrow$4.62}}  &   {\color{Peach}{$\uparrow$6.00}}  
 &   {\color{Peach}{$\uparrow$5.44}}  &   {\color{Peach}{$\uparrow$6.28}}  
&   {\color{Peach}{$\uparrow$7.88}}  &  & 
    {\color{Peach}{$\downarrow$0.08}}   & {\color{Peach}{$\downarrow$0.07}} &{\color{Peach}{$\downarrow$0.27}} & {\color{Peach}{$\downarrow$0.37}} & {\color{Peach}{$\downarrow$0.23}}  & {\color{Peach}{$\downarrow$0.44}}& {\color{Peach}{$\downarrow$0.69}}
\\
\bottomrule
\end{tabular}
}
\end{table*}

\noindent{\bf Implementation Details.} Considering the unified timestep policy, also to make generation and diffusion-classifier settings directly comparable, we apply the same token-reduction policy at every denoising step $t$. We do not cache merged pairings or pooling indices across timesteps, all reductions are recomputed per step and per block. This avoids $t$-dependent artifacts for synthesis and, because the diffusion classifier is a Monte-Carlo estimator over ($t$, $\epsilon$), keeps the schedule temporally consistent, reducing variance.  

\subsection{Comparisons to State-of-the-Art Approaches}
\label{sec:cls-exps}

Table 1 presents the comparisons with state-of-the-art approaches on Oxford-IIIT Pets using Top-1 accuracy under $\sim$10\% FLOPs reduction. The no-acceleration baseline is 81.03\%. Token-merging/pruning baselines suffer large drops: ToMe (8.07\%) and SiTo (12.19\%), with pruning methods DiP-GO (4.50\%) and MosaicDiff ({3.65\%}), showing that compression tuned for synthesis often harms recognition. Our Laplacian-gated merging ({\bf BiGain$_\texttt{TM}$}) retains far more accuracy (78.38\%, 2.65\% drop), cutting the loss by 27$\sim$78\% vs. these methods at matched FLOPs. In the downsampling regime (14.2\% FLOPs), ToDo slightly decreases the accuracy (-1.88\%), while our Interpolate-Extrapolate KV-downsampling ({\bf BiGain$_\texttt{TD}$}) is the best overall (79.90\%, only 1.13\% drop), also with much better generation ability than ToDo, as we will discuss later. Overall, {\bf BiGain} delivers the strongest classification under comparable compute.

\subsection{Classification vs. Generation}

\begin{figure}[t]
  \centering
  \includegraphics[width=\linewidth]{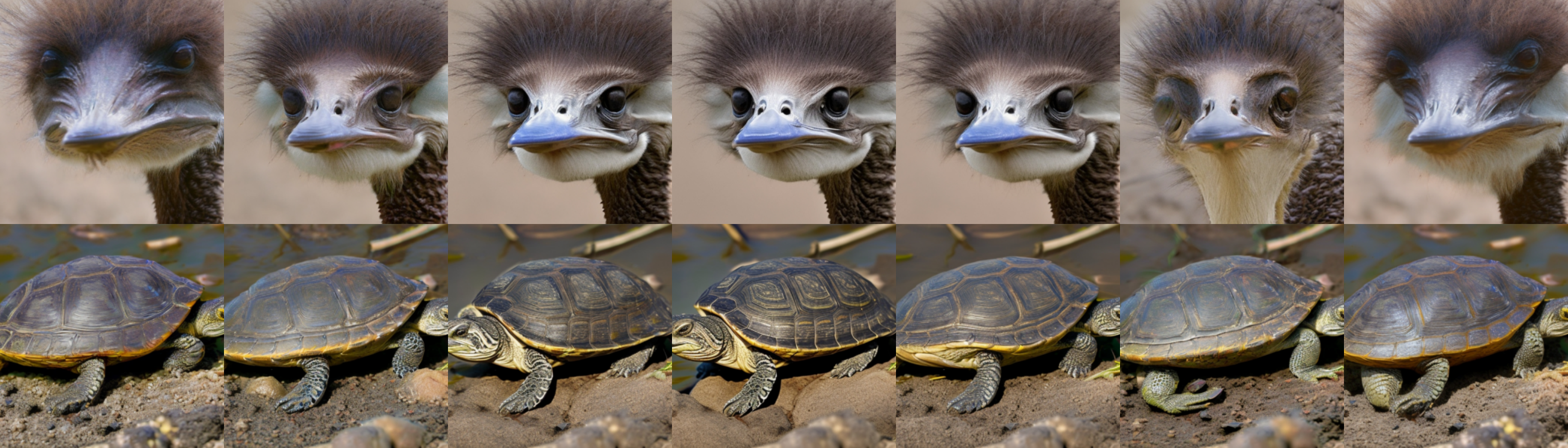}
  \vspace{-1.6em}
  \caption{Qualitative comparison of BiGain$_\texttt{TD}$ and ToDo on SD-2.0 backbone. From left to right: BiGain$_\texttt{TD}$ with downsampling factors $4\times$, $3\times$, and $2\times$, no acceleration, then ToDo with factors $2\times$, $3\times$, and $4\times$.}
  \label{fig:todo_comparison}
\vspace{-1.2em}
\end{figure}

We provide classification and generation comparisons under {\em Token Downsampling} in Table~\ref{tab:sd20-down_classification} (SD-2.0 backbone) and Table~\ref{tab:dit-down_classification} (DiT-XL/2). As shown, our method consistently outperforms the baseline, and the advantage becomes more pronounced as the downsampling ratio increases. The same trend holds for generation: with higher downsampling factors, our approach yields increasingly better results. We further observe (Table~\ref{tab:dit-down_classification}) that the ToDo method performs very poorly on the DiT-XL/2 model, whereas our method remains more stable on this backbone. Furthermore, with relatively small downsampling (2$\times$), our method can match or slightly surpass the original unaccelerated model in both classification and generation (see Fig.~\ref{fig:todo_comparison} for a qualitative comparison in generation task).

\begin{figure}[t]
  \centering
  \includegraphics[width=\linewidth]{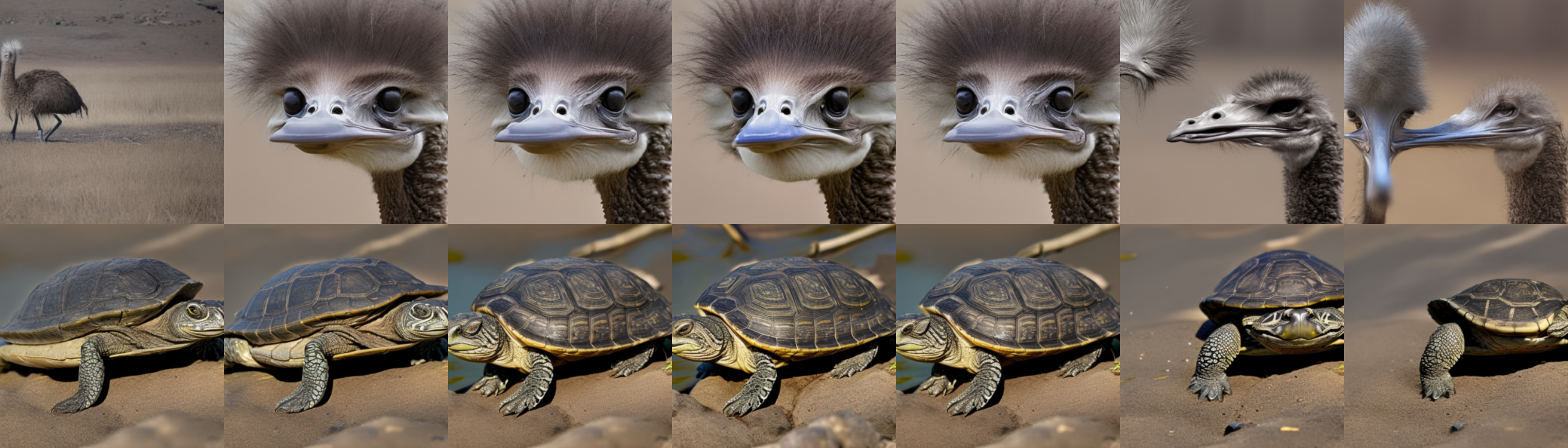}
  \vspace{-1.6em}
  \caption{Qualitative comparison of BiGain$_\texttt{TM}$ and ToMe on SD-2.0 backbone. From left to right: BiGain$_\texttt{TM}$ with 70\%, 50\%, and 30\% merge ratios, no acceleration, then ToMe with 30\%, 50\%, and 70\% merge ratios.}
  \label{fig:tome_comparison}
\vspace{-.8em}
\end{figure}

For {\em Token Merging}, classification and generation comparisons are provided in Table~\ref{tab:sd20-merge_classification} (SD-2.0 backbone) and Table~\ref{tab:dit_merge_cls_gen} (DiT-XL/2). The results mirror those under downsampling: as the merging ratio increases (i.e., with more aggressive pruning), our method achieves substantially better performance than the baseline. In particular, our classification accuracy significantly surpasses ToMe, while our generation capability also exceeds it (see Fig.~\ref{fig:tome_comparison} for a qualitative comparison). These results highlight the dual advantages of our approach in both classification and generation.

\begin{table*}[h]
\centering
\caption{\textbf{Ablation of token-merging locations in Stable Diffusion~2.0 on Pets.}
Self-Attention (SA) is always merged; Cross-Attention (CA) and MLP are toggled. Results reported at merge ratios $r\!\in\!\{0.7,0.5,0.3\}$. The underlined results indicate the best performance across all configurations.}
\label{tab:pets-ablation-modules}
\newcommand{\best}[1]{\textbf{#1}}
\resizebox{.78\textwidth}{!}{
\begin{tabular}{lcccccccccccc}
\toprule
& \multicolumn{3}{c}{\textbf{SA only}} &
  \multicolumn{3}{c}{\textbf{SA+CA}} &
  \multicolumn{3}{c}{\textbf{SA+MLP}} &
  \multicolumn{3}{c}{\textbf{SA+CA+MLP}} \\
\cmidrule(lr){2-4} \cmidrule(lr){5-7} \cmidrule(lr){8-10} \cmidrule(lr){11-13}
\textbf{Method} & \textbf{0.7} & \textbf{0.5} & \textbf{0.3} & \textbf{0.7} & \textbf{0.5} & \textbf{0.3} & \textbf{0.7} & \textbf{0.5} & \textbf{0.3} & \textbf{0.7} & \textbf{0.5} & \textbf{0.3} \\
\midrule
ToMe \cite{bolya2023token}  & 65.76& 72.96 &  78.44   & 61.68 & 68.41 & 74.46 & 51.43 & 58.71 & 66.35 & 50.86 & 59.53 & 66.20 \\
\textbf{BiGain$_\texttt{TM}$ (Ours)} & \cellcolor{lightgray!50}\best{\uline{74.63}}
& \cellcolor{lightgray!50}\best{\uline{78.38}}
& \cellcolor{lightgray!50}\best{\uline{80.40}}
& \cellcolor{lightgray!50}\bf 73.89
& \cellcolor{lightgray!50}\bf 78.03
& \cellcolor{lightgray!50}\bf 79.56
& \cellcolor{lightgray!50}\bf 68.27
& \cellcolor{lightgray!50}\bf 74.93
& \cellcolor{lightgray!50}\bf 77.98
& \cellcolor{lightgray!50}\bf 68.25
& \cellcolor{lightgray!50}\bf 74.84
& \cellcolor{lightgray!50}\bf 77.95 \\
\bottomrule
\end{tabular}
}
\end{table*}

\begin{table*}[t]
\centering
\caption{Further speedup on SD-2.0 \textbf{Token Merging}: classification performance vs. merge ratio. Acc@1 for single-label datasets; Acc@1 and mAP for multi-label COCO-2017. GFLOPs are measured at merge ratio r = 0.7.}
\label{tab:sd20-merge_classification_ana}
\vspace{-0.1in}
\small
\setlength{\tabcolsep}{5pt}
\resizebox{.9\textwidth}{!}{
\begin{tabular}{llcccccccc}
\toprule
Dataset & Method & GFLOPs  & 10\% & 20\% & 30\% & 40\% & 50\% & 60\% & 70\% \\
\midrule
\multirow{3}{*}{Pets} 
  & Laplacian Gated Merge & {704.99} & {81.16} & {81.16} & {80.4} & {80.07} & {78.38} & {76.04} & {74.63} \\
  & Cached Assignment Merge & {698.88} & {80.29} & {79.97} & {79.89} & {79.01} & {78.11} & {75.91} & {74.49} \\
  & Adaptive Block Merge & {695.08} & {80.40} & {80.16} & {79.99} & {79.18} & {77.84} & {75.96} & {74.13} \\
\midrule
\multirow{3}{*}{ImageNet-100} 
  &  Laplacian Gated Merge & {704.99} & {71.94} & {71.62} & {70.64} & {68.38} & {67.24} & {65.20} & {61.28}  \\
  &  Cached Assignment Merge & {698.88} & {71.76} & {71.16} & {70.44} & {69.38} & { 67.78 } & {64.56} & {61.28} \\
  &  Adaptive Block Merge & {695.08} & {72.58} & {71.94} & {70.58} & {70.52} & {68.04} & {65.36} & {60.98} \\
\midrule
\multirow{3}{*}{ImageNet-1K} 
  &  Laplacian Gated Merge & {704.99}& {57.25} & {56.50} & {55.80 } & {54.80} & {52.50} & {49.10} & {44.50 }  \\
  &  Cached Assignment Merge & {698.88} & { 56.30} & {56.05 } & {56.05} & {53.15} & {52.30 } & {47.90} & {44.60 } \\
  &  Adaptive Block Merge & {695.08} & {56.95} & {56.25} & {56.00} & {54.60} & {51.95} & {48.20} & {44.85} \\
\midrule
\multirow{6}{*}{COCO-2017}
  & \emph{Acc@1} $|$ Laplacian Gated Merge & 704.99& {71.96} & {71.56} & {70.64} & {69.2} & {67.64} & {64.94} & {61.44} \\
  & \emph{Acc@1} $|$ Cached Assignment Merge & 698.88 & {71.72} & {71.40} & {70.22} & {70.02} & {67.94 } & {64.88} & {60.88} \\
  & \emph{Acc@1} $|$ Adaptive Block Merge & 695.08 & {71.76} & {71.44} & {70.28} & {69.62} & {67.26} & {64.70} & {60.56} \\
\cmidrule(lr){2-10}
  & \emph{mAP} $|$ Laplacian Gated Merge   & 704.99 & {46.38} & {46.21} & {46.05} & {45.50} & {44.94} & {43.98} & {42.44} \\
  & \emph{mAP} $|$ Cached Assignment Merge   & 698.88 & {46.30} & {46.32} & {45.94} & {45.96} & {45.19} & {43.93} & {42.41} \\
  & \emph{mAP} $|$ Adaptive Block Merge & 695.08 & {46.35} & {46.41} & {45.93} & {45.87} & {45.12} & {43.96} & {42.32} \\
\bottomrule
\end{tabular}
}
\end{table*}

\subsection{Ablation}
\label{sec:ablation}

\noindent{\bf Where to reduce}\footnote{Unless noted otherwise, ablations are conducted on \texttt{Oxford-IIIT Pets} with identical sampling schedules, classifier settings (for classification ablations), and reduction ratios as in the main results.}. As shown in Table~\ref{tab:pets-ablation-modules}, we compare applying token reduction to \emph{self-attention} {\em only} (SA), \emph{self+cross attentions} (SA+CA), and \emph{self+cross+MLP} (SA+CA+MLP). We find that \textbf{SA-only} consistently delivers the best quality-efficiency trade-off: it preserves prompt adherence (avoiding CA degradation) and avoids compounding bias through MLP compression. On Pets, SA-only attains the highest accuracy, while SA+MLP reduces prompt fidelity and SA+CA+MLP further harms fine details. \emph{Conclusion:} we adopt \textbf{SA-only} reduction as default for all SD~2.0 experiments.

\begin{table}
\centering
\caption{\textbf{Ablation over token scoring heuristics for Stable Diffusion 2.0.}
Top-1 acc. (\%) on Pets dataset across merge ratios. Local Laplacian signals outperform global or spectral metrics.}
\label{tab:pets-ablation-merge}
\small
\setlength{\tabcolsep}{8pt}
\begin{tabular}{lccc}
\toprule
\textbf{Scoring method} & \textbf{0.7} & \textbf{0.5} & \textbf{0.3} \\ \midrule
Global mean deviation & 72.96 & 77.84  & 79.91  \\
$\ell_1$-norm               & 73.02 & 77.11  & 79.86 \\
$\ell_2$-norm               & 72.72 & 77.95  & 79.61 \\
Channel variance      & 73.04 & 77.95  & 79.83 \\
Laplacian Filter $\ell_1$   & \bf{74.63} & \bf{78.38}  & \bf{80.40} \\
Laplacian Filter $\ell_2$   & 74.24 & 77.81  & 79.80 \\
DFT spectral centroid & 73.75 & 77.92  & 79.10 \\
DFT amplitude         & 73.10 & 77.76  & 79.34 \\
Cosine to neighbors   & 74.00 & 78.22  & 79.56 \\
Cosine to global mean & 73.32 & 77.84  & 79.83 \\
\bottomrule
\end{tabular}
\vspace{-0.07in}
\end{table}

\noindent{\bf How to score tokens}. As shown in Table~\ref{tab:pets-ablation-merge}, local frequency cues dominate: Laplacian Filter ($\ell_1$) is best at all merge ratios, outperforming global statistics (norms, channel variance), spectral DFT measures, and cosine similarity by 0.3$\sim$1.9\%. This supports our frequency-aware design and motivates using a Laplacian proxy for gated merging. Overall, for SD-2.0, token merging in SA with Laplacian scoring provides the strongest quality-efficiency trade-off under our ablation protocol. The detailed mathematical formulations of these score heuristics can be seen in supplementary material.

\vspace{-0.05in}
\subsection{Analysis}

\begin{figure}[t] 
  \centering
  \includegraphics[width=\linewidth]{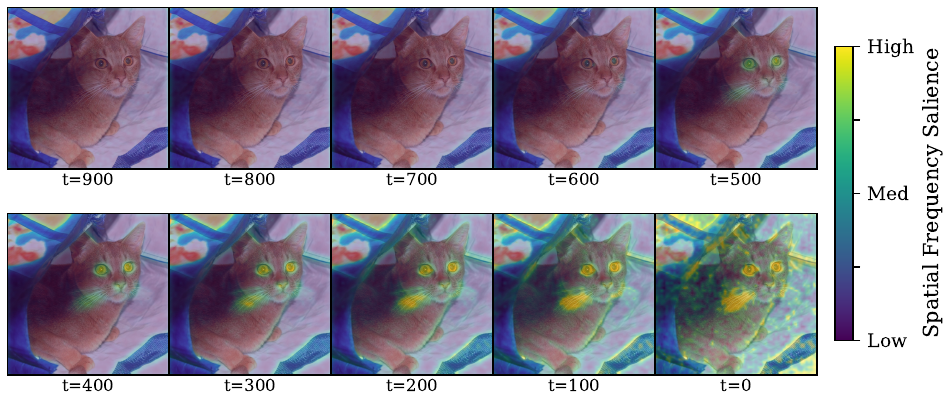}
  \vspace{-0.27in}
    \caption{Visualization of our Laplacian-based frequency heuristic on hidden 
    representations from Stable Diffusion-2.0. We probe U-Net at the 
    highest-resolution upsampling stage. 
    The visualization is computed from a noised image without a text prompt, 
    showing the model's intrinsic frequency-aware reconstruction dynamics. 
    To reduce variance, we randomly sample 100 independent noise realizations 
    and visualize the averaged token salience map.}
  \label{fig:token-importance_vis1}
   \vspace{-0.13in}
\end{figure}

\noindent{\bf Further Speedup.} {\em Our vanilla Laplacian Merge.} Before the Q/K/V projections, we run a 2-D Laplacian filter on the hidden map to score each token by local frequency (contrast w.r.t. its four neighbors). We then partition the feature map into $s_x\times s_y$ cells; within each cell, low-frequency tokens serve as \emph{destinations} and the remaining \emph{source} tokens are greedily assigned by cosine similarity. Because merging acts like a low-pass filter that can destroy high-freq detail, we restrict merging to low-freq tokens only. {\em Two faster variants.} (1) {\em Our Cached Assignment Merge}: in the highest-resolution U-Net stages (two Transformer blocks for down sampling and three for up sampling), compute the merge/unmerge map once in the first attention block and reuse it within the stage. (2) {\em Our Adaptive Block Merge}: after computing Laplacian scores, aggregate them per cell and merge entire low-frequency cells with no per-token matching, yielding extra speed with minimal accuracy loss. As shown in Table~\ref{tab:sd20-merge_classification_ana}, both variants closely track Laplacian-Gated Merge across 10$\sim$70\% merge ratios across different datasets while providing additional FLOPs savings.

\begin{figure}[h]
  \centering
  \includegraphics[width=0.95\linewidth]{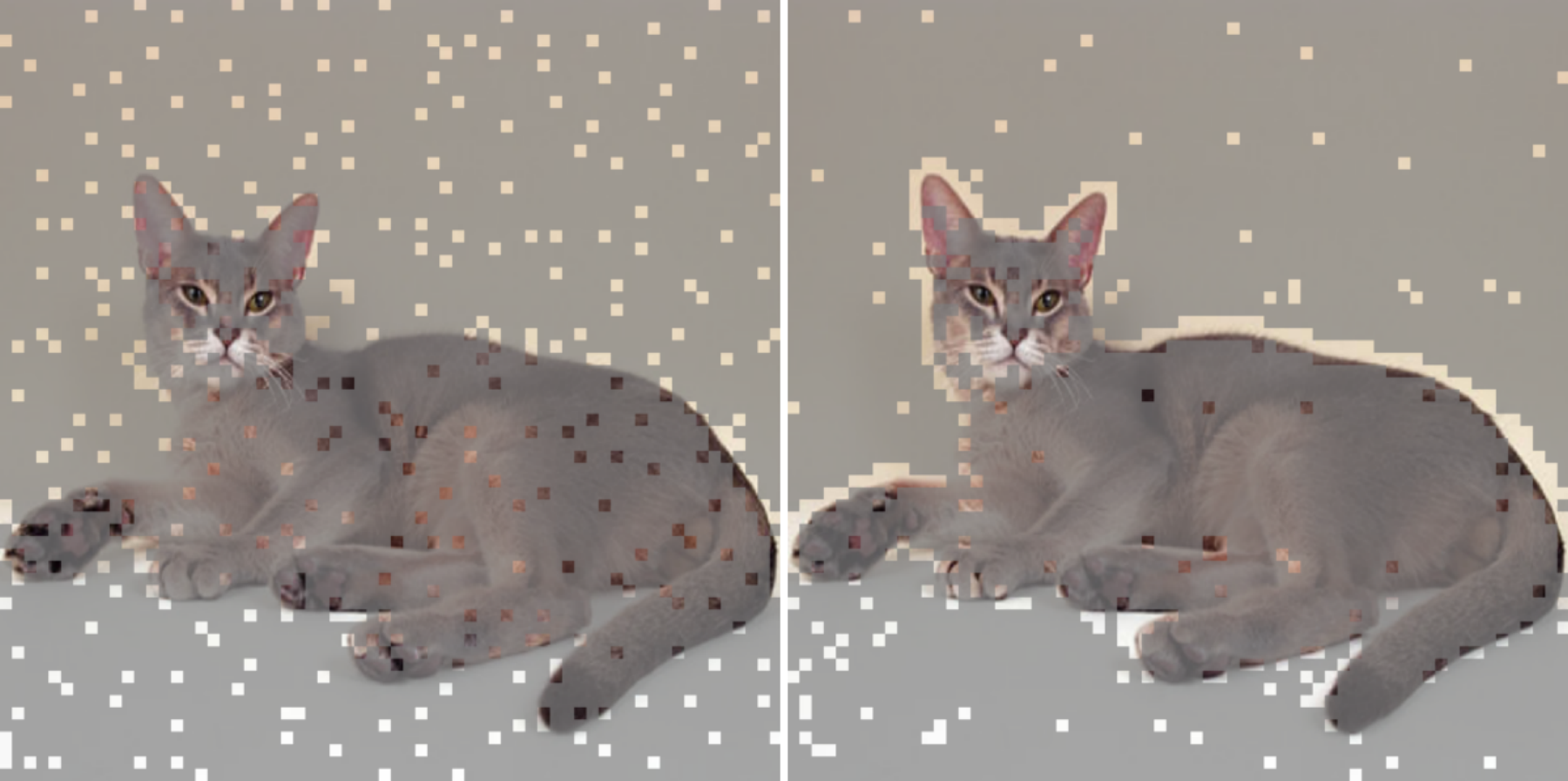}
  \vspace{-0.1in}
  \caption{\textbf{Comparison of token merging schemes.} 
    Left: ToMe~\cite{bolya2023token}; Right: Our {\bf BiGain$_\texttt{TM}$}. 
    Merging is applied with a merge ratio 90\% at the highest-resolution latent layer of the U-Net transformer in Stable Diffusion~2.0 at denoising step $t=200$. 
    Grayscale indicates merged tokens. }
  \label{fig:merge_scheme_vis2}
  \vspace{-0.1in}
\end{figure}

\noindent{\bf Visualization.}
We compare token-importance maps for generation and classification to reveal their different spectral needs. As in Fig.~\ref{fig:token-importance_vis1}, frequency-aware reduction yields a favorable bias-variance trade-off: retaining low-frequency tokens stabilizes classification, while selectively keeping high-frequency tokens preserves generation quality, making one heuristic effective for both tasks. To illustrate our Laplacian scoring, we probe SD-2.0 at the highest-resolution upsampling block and visualize pre-attention hidden states filtered by a 2-D Laplacian. Maps are averaged over 100 noise draws without a text prompt to reduce variance, to reveal model's intrinsic frequency sensitivity.

In Fig.~\ref{fig:merge_scheme_vis2}, we compare ToMe {\em vs.} \textbf{BiGain\textsubscript{TM}} at 90\% merge on the highest-resolution U-Net transformer layer at $t=$200 (grayscale = merged). Laplacian-gated merging preserves more class-discriminative structure (e.g., the cat's edges) than standard ToMe.

\vspace{-0.05in}
\section{Conclusion}
\label{sec:conclusion}
\vspace{-0.05in}

In this work, we revisited token compression for diffusion models as a bi-objective problem, preserving both generative and discriminative abilities, and introduced {\bf BiGain}, a training-free framework built on two frequency-aware operators: {\em Laplacian-Gated Token Merging} (merge in smooth regions, keep edges) and {\em Interpolate-Extrapolate KV-Downsampling} (downsample K/V with controllable interextrapolation while keeping Q unchanged). Using DiT/U-Net backbones and multiple datasets, BiGain consistently improves the speed-accuracy trade-off for diffusion-based classification while maintaining, and sometimes even improving generation quality under comparable compute.

\section*{Acknowledgements}

This work was supported by the MBZUAI–WIS Joint Program for AI Research.

{
    \small
    \bibliographystyle{ieeenat_fullname}
    \bibliography{main}

@String(ICCV= {Int. Conf. Comput. Vis.})

@String(ICLR = {Int. Conf. Learn. Represent.})

@String(AAAI = {AAAI})

@String(ICCV  = {ICCV})

@String(ICLR  = {ICLR})

@article{song2020denoising,
  title={Denoising diffusion implicit models},
  author={Song, Jiaming and Meng, Chenlin and Ermon, Stefano},
  journal={arXiv preprint arXiv:2010.02502},
  year={2020}
}

@article{smith2024todo,
  title={Todo: Token downsampling for efficient generation of high-resolution images},
  author={Smith, Ethan and Saxena, Nayan and Saha, Aninda},
  journal={arXiv preprint arXiv:2402.13573},
  year={2024}
}

@inproceedings{li2023your,
  title={Your diffusion model is secretly a zero-shot classifier},
  author={Li, Alexander C and Prabhudesai, Mihir and Duggal, Shivam and Brown, Ellis and Pathak, Deepak},
  booktitle={Proceedings of the IEEE/CVF International Conference on Computer Vision},
  pages={2206--2217},
  year={2023}
}

@article{clark2023text,
  title={Text-to-image diffusion models are zero shot classifiers},
  author={Clark, Kevin and Jaini, Priyank},
  journal={Advances in Neural Information Processing Systems},
  volume={36},
  pages={58921--58937},
  year={2023}
}

@inproceedings{rombach2022high,
  title={High-resolution image synthesis with latent diffusion models},
  author={Rombach, Robin and Blattmann, Andreas and Lorenz, Dominik and Esser, Patrick and Ommer, Bj{\"o}rn},
  booktitle={Proceedings of the IEEE/CVF conference on computer vision and pattern recognition},
  pages={10684--10695},
  year={2022}
}

@inproceedings{peebles2023scalable,
  title={Scalable diffusion models with transformers},
  author={Peebles, William and Xie, Saining},
  booktitle={Proceedings of the IEEE/CVF international conference on computer vision},
  pages={4195--4205},
  year={2023}
}

@inproceedings{parkhi2012cats,
  title={Cats and dogs},
  author={Parkhi, Omkar M and Vedaldi, Andrea and Zisserman, Andrew and Jawahar, CV},
  booktitle={2012 IEEE conference on computer vision and pattern recognition},
  pages={3498--3505},
  year={2012},
  organization={IEEE}
}

@inproceedings{tian2020contrastive,
  title={Contrastive multiview coding},
  author={Tian, Yonglong and Krishnan, Dilip and Isola, Phillip},
  booktitle={European conference on computer vision},
  pages={776--794},
  year={2020},
  organization={Springer}
}

@inproceedings{lin2014microsoft,
  title={Microsoft coco: Common objects in context},
  author={Lin, Tsung-Yi and Maire, Michael and Belongie, Serge and Hays, James and Perona, Pietro and Ramanan, Deva and Doll{\'a}r, Piotr and Zitnick, C Lawrence},
  booktitle={European conference on computer vision},
  pages={740--755},
  year={2014},
  organization={Springer}
}

@article{lu2022dpm,
  title={Dpm-solver: A fast ode solver for diffusion probabilistic model sampling in around 10 steps},
  author={Lu, Cheng and Zhou, Yuhao and Bao, Fan and Chen, Jianfei and Li, Chongxuan and Zhu, Jun},
  journal={Advances in neural information processing systems},
  volume={35},
  pages={5775--5787},
  year={2022}
}

@inproceedings{salimansprogressive,
  title={Progressive Distillation for Fast Sampling of Diffusion Models},
  author={Salimans, Tim and Ho, Jonathan},
  booktitle={International Conference on Learning Representations},
  year={2022}
}

@article{ryoo2021tokenlearner,
  title={Tokenlearner: Adaptive space-time tokenization for videos},
  author={Ryoo, Michael and Piergiovanni, AJ and Arnab, Anurag and Dehghani, Mostafa and Angelova, Anelia},
  journal={Advances in neural information processing systems},
  volume={34},
  pages={12786--12797},
  year={2021}
}

@inproceedings{bolya2023token,
  title={Token Merging: Your ViT But Faster},
  author={Bolya, Daniel and Fu, Cheng-Yang and Dai, Xiaoliang and Zhang, Peizhao and Feichtenhofer, Christoph and Hoffman, Judy},
  booktitle={ICLR},
  year={2023}
}

@article{russakovsky2015imagenet,
  title={Imagenet large scale visual recognition challenge},
  author={Russakovsky, Olga and Deng, Jia and Su, Hao and Krause, Jonathan and Satheesh, Sanjeev and Ma, Sean and Huang, Zhiheng and Karpathy, Andrej and Khosla, Aditya and Bernstein, Michael and others},
  journal={International journal of computer vision},
  year={2015},
  publisher={Springer}
}

@article{chen2023robust,
  title={Robust classification via a single diffusion model},
  author={Chen, Huanran and Dong, Yinpeng and Wang, Zhengyi and Yang, Xiao and Duan, Chengqi and Su, Hang and Zhu, Jun},
  journal={arXiv preprint arXiv:2305.15241},
  year={2023}
}

@article{ho2020denoising,
  title={Denoising diffusion probabilistic models},
  author={Ho, Jonathan and Jain, Ajay and Abbeel, Pieter},
  journal={Advances in neural information processing systems},
  year={2020}
}

@article{labs2025flux1kontextflowmatching,
  title={Flux. 1 kontext: Flow matching for in-context image generation and editing in latent space},
  author={Batifol, Stephen and Blattmann, Andreas and Boesel, Frederic and Consul, Saksham and Diagne, Cyril and Dockhorn, Tim and English, Jack and English, Zion and Esser, Patrick and Kulal, Sumith and others},
  journal={arXiv e-prints},
  pages={arXiv--2506},
  year={2025}
}

@inproceedings{fang2023structural,
  title={Structural pruning for diffusion models},
  author={Gongfan Fang and Xinyin Ma and Xinchao Wang},
  booktitle={Advances in Neural Information Processing Systems},
  year={2023},
}

@article{zhu2024dip,
  title={Dip-go: A diffusion pruner via few-step gradient optimization},
  author={Zhu, Haowei and Tang, Dehua and Liu, Ji and Lu, Mingjie and Zheng, Jintu and Peng, Jinzhang and Li, Dong and Wang, Yu and Jiang, Fan and Tian, Lu and others},
  journal={Advances in Neural Information Processing Systems},
  year={2024}
}

@inproceedings{castells2024ld,
  title={Ld-pruner: Efficient pruning of latent diffusion models using task-agnostic insights},
  author={Castells, Thibault and Song, Hyoung-Kyu and Kim, Bo-Kyeong and Choi, Shinkook},
  booktitle={Proceedings of the IEEE/CVF Conference on Computer Vision and Pattern Recognition Workshop},
  year={2024}
}

@inproceedings{zhang2025training,
  title={Training-free and hardware-friendly acceleration for diffusion models via similarity-based token pruning},
  author={Zhang, Evelyn and Tang, Jiayi and Ning, Xuefei and Zhang, Linfeng},
  booktitle={Proceedings of the AAAI Conference on Artificial Intelligence},
  year={2025}
}

@article{Peebles2022DiT,
  title={Scalable Diffusion Models with Transformers},
  author={William Peebles and Saining Xie},
  year={2022},
  journal={arXiv preprint arXiv:2212.09748},
}

@inproceedings{guoiccv2025,
  title={MosaicDiff: Training-free Structural Pruning for Diffusion Model Acceleration Reflecting Pretraining Dynamics},
  author={Bowei Guo and Shengkun Tang and Cong Zeng and Zhiqiang Shen},
  booktitle={ICCV},
  year={2025}
}

@article{paszke2019pytorchimperativestylehighperformance,
  title={Pytorch: An imperative style, high-performance deep learning library},
  author={Paszke, Adam and Gross, Sam and Massa, Francisco and Lerer, Adam and Bradbury, James and Chanan, Gregory and Killeen, Trevor and Lin, Zeming and Gimelshein, Natalia and Antiga, Luca and others},
  journal={Advances in neural information processing systems},
  volume={32},
  year={2019}
}

@article{meng2024not,
  title={Not all diffusion model activations have been evaluated as discriminative features},
  author={Meng, Benyuan and Xu, Qianqian and Wang, Zitai and Cao, Xiaochun and Huang, Qingming},
  journal={Advances in Neural Information Processing Systems},
  volume={37},
  pages={55141--55177},
  year={2024}
}

@article{tang2023emergent,
  title={Emergent correspondence from image diffusion},
  author={Tang, Luming and Jia, Menglin and Wang, Qianqian and Phoo, Cheng Perng and Hariharan, Bharath},
  journal={Advances in Neural Information Processing Systems},
  volume={36},
  pages={1363--1389},
  year={2023}
}

@article{Seitzer2020FID,
  author  = {Seitzer, Maximilian},
  title   = {{pytorch-fid}: {FID} Score for {PyTorch}},
  journal = {GitHub repository},
  year    = {2023},
  url     = {https://github.com/mseitzer/pytorch-fid},
  note    = {Version 0.3.0},
}

@article{favero2025conditional,
  title={Conditional diffusion models are medical image classifiers that provide explainability and uncertainty for free},
  author={Favero, Gian Mario and Saremi, Parham and Kaczmarek, Emily and Nichyporuk, Brennan and Arbel, Tal},
  journal={arXiv preprint arXiv:2502.03687},
  year={2025}
}

@inproceedings{he2024diffusion,
  title={A diffusion-based framework for multi-class anomaly detection},
  author={He, Haoyang and Zhang, Jiangning and Chen, Hongxu and Chen, Xuhai and Li, Zhishan and Chen, Xu and Wang, Yabiao and Wang, Chengjie and Xie, Lei},
  booktitle={Proceedings of the AAAI conference on artificial intelligence},
  volume={38},
  number={8},
  pages={8472--8480},
  year={2024}
}

@inproceedings{beizaee2025correcting,
  title={Correcting deviations from normality: A reformulated diffusion model for multi-class unsupervised anomaly detection},
  author={Beizaee, Farzad and Lodygensky, Gregory A and Desrosiers, Christian and Dolz, Jose},
  booktitle={Proceedings of the Computer Vision and Pattern Recognition Conference},
  pages={19088--19097},
  year={2025}
}

@article{sousa2025data,
  title={Data augmentation in earth observation: A diffusion model approach},
  author={Sousa, Tiago and Ries, Beno{\^\i}t and Guelfi, Nicolas},
  journal={Information},
  volume={16},
  number={2},
  pages={81},
  year={2025},
  publisher={MDPI}
}

@inproceedings{bolya2023tokenmergingfaststable,
  title={Token merging for fast stable diffusion},
  author={Bolya, Daniel and Hoffman, Judy},
  booktitle={Proceedings of the IEEE/CVF conference on computer vision and pattern recognition},
  pages={4599--4603},
  year={2023}
}
}

\clearpage

\appendix

\section*{\Large{Supplementary Material}}

\etocdepthtag.toc{mtappendix}
\etocsettagdepth{mtchapter}{none}
\etocsettagdepth{mtappendix}{subsection}  
\tableofcontents

\newpage

\section{Frequency-Aware Token Reduction and Diffusion Classification}
\label{sec:theory}

In this section we provide a simple stylized model showing how frequency-aware token reduction can tighten the diffusion-classifier decision, viewed through a margin--variance trade-off on the paired Monte Carlo estimator introduced in Sec.~3.1.1 of the main paper.

\subsection{Diffusion Classifier and Paired Difference}

Recall the diffusion-classifier score \cite{li2023your}:
\begin{equation}
\begin{aligned}
S(x,c)
&\;=\;
\E_{t,\epsilon}
\left\|
\epsilon - \epsilon_\theta(x_t,c,t)
\right\|_2^2,\\
x_t
&=
\sqrt{\bar\alpha_t}\,x
+
\sqrt{1-\bar\alpha_t}\,\epsilon.
\end{aligned}
\label{eq:dc-score}
\end{equation}
The prediction is $\hat c(x)=\arg\min_{c \in \mathcal{C}} S(x,c)$, implemented using \emph{paired sampling}: for a shared Monte Carlo set $\mathcal{S}_{\mathrm{MC}}=\{(t_s,\epsilon_s)\}_{s=1}^{S_{\mathrm{MC}}}$ (reused for all classes as in Sec.~\ref{D_classifier}) we approximate $S(x,c)$ by
\begin{equation}
\begin{aligned}
\hat S_{S_{\mathrm{MC}}}(x,c)
&=
\frac{1}{S_{\mathrm{MC}}}\sum_{s=1}^{S_{\mathrm{MC}}}
\ell(x,c;t_s,\epsilon_s),\\
\ell(x,c;t,\epsilon)
&=
\left\|
\epsilon - \epsilon_\theta(x_t,c,t)
\right\|_2^2.
\end{aligned}
\label{eq:paired-score}
\end{equation}

Fix the true class $c^\star$ and a distractor $\tilde c$. For one paired draw $(t,\epsilon)\sim p(t)p(\epsilon)$, we define
\begin{equation}
D(t,\epsilon)
=
\ell(x,\tilde c;t,\epsilon)
-
\ell(x,c^\star;t,\epsilon).
\label{eq:paired-diff}
\end{equation}
Let
\begin{equation}
\begin{aligned}
\mu
&=
\E[D],
\sigma^2=\Var[D].
\end{aligned}
\end{equation}
With $S_{\mathrm{MC}}$ shared samples we obtain the paired estimator
\begin{equation}
\begin{aligned}
\widehat{\Delta}_{S_{\mathrm{MC}}}
&=
\frac{1}{S_{\mathrm{MC}}}\sum_{s=1}^{S_{\mathrm{MC}}} D(t_s,\epsilon_s),\\
\E[\widehat{\Delta}_{S_{\mathrm{MC}}}]
&=
\mu,
\Var(\widehat{\Delta}_{S_{\mathrm{MC}}})=
\frac{\sigma^2}{S_{\mathrm{MC}}}.
\end{aligned}
\label{eq:delta-hat}
\end{equation}
For a consistent classifier we have $\mu>0$, and misclassification against $\tilde c$ corresponds to the tail event $\widehat{\Delta}_{S_{\mathrm{MC}}} \le 0$.

By Cantelli’s inequality, for any random variable $Z$ with mean $m$ and variance $v$, \( \Pr(Z-m \le -a) \le v/(v+a^2)\) for $a>0$. Applying this to $Z=\widehat{\Delta}_{S_{\mathrm{MC}}}$, $m=\mu$, $a=\mu$ yields
\begin{equation}
\begin{aligned}
\Pr\bigl(\widehat{\Delta}_{S_{\mathrm{MC}}} \le 0\bigr)
&\;\le\;
\frac{\sigma^2/S_{\mathrm{MC}}}{\mu^2 + \sigma^2/S_{\mathrm{MC}}}
=
f(r),
r=
\frac{\sigma}{\mu},
\end{aligned}
\label{eq:cantelli}
\end{equation}
where $f(r)$ is strictly increasing in $r$. Thus, tightening the Cantelli bound is equivalent to decreasing the ratio $r$.

\subsection{Bandwise Decomposition}

We now introduce a simple frequency-domain model of the paired difference. Let $\{\phi_k\}$ be an orthonormal 2-D DCT/Fourier basis over the spatial token grid. For each $(t,\epsilon)$ and class $c$, we expand the error as:
\begin{equation}
\ell(x,c;t,\epsilon)
=
\sum_k \omega_k(t)\,
\bigl|
\widehat{\epsilon}(k)
-
\widehat{\epsilon_\theta}(k;t,c)
\bigr|^2,
\label{eq:band-decomp}
\end{equation}
where $\widehat{\cdot}(k)$ denotes the coefficient in band $k$ and $\omega_k(t)\ge 0$ is a per-band reliability weight (e.g., arising from the ELBO weighting over $t$). 

We define the bandwise paired difference as:
\begin{equation}
\Delta_k(t,\epsilon)
:=
\bigl|
\widehat{\epsilon}(k)
-
\widehat{\epsilon_\theta}(k;t,\tilde c)
\bigr|^2
-
\bigl|
\widehat{\epsilon}(k)
-
\widehat{\epsilon_\theta}(k;t,c^\star)
\bigr|^2,
\label{eq:delta-k}
\end{equation}
so that
\begin{equation}
D(t,\epsilon)
=
\sum_k \omega_k(t)\,\Delta_k(t,\epsilon).
\end{equation}
Let
\begin{equation}
\begin{aligned}
\mu_k
&:=
\E_{t,\epsilon}[\Delta_k(t,\epsilon)],\\
\sigma_k^2
&:=
\Var_{t,\epsilon}[\Delta_k(t,\epsilon)],\\
w_k
&:=
\E_t[\omega_k(t)].
\end{aligned}
\label{eq:band-moments}
\end{equation}
Assuming that cross-band covariances are weak,
\begin{equation}
\Cov\bigl(\Delta_i,\Delta_j\bigr)\approx 0,
\qquad
i\neq j,
\label{eq:weak-cov}
\end{equation}
we obtain the approximation
\begin{equation}
\begin{aligned}
\mu
&\approx
\sum_k w_k \mu_k,\\
\sigma^2
&\approx
\sum_k w_k^2 \sigma_k^2.
\end{aligned}
\label{eq:mu-sigma-band}
\end{equation}

Intuitively, discriminative classifiers benefit from high-frequency components (edges and textures) to refine the class margin~$\mu$. Coarse structures (low frequencies) provide stable semantic cues, while fine structures (high frequencies) often distinguish specific classes. Consequently, although high-frequency bands may exhibit larger variance under stochastic sampling, they can still carry strong class-specific information for the correct class. A reduction strategy that indiscriminately suppresses these frequencies, effectively acting as a low-pass filter with $H_P(k) \approx 0$ for high~$k$, risks inducing a large margin loss $\Delta\mu$ that can outweigh any variance reduction $\Delta\sigma^2$.

\subsection{Token Reduction as a Spectral Operator}

We consider an attention block operating on a window of tokens, with pre-reduction features $z_i = s_i + n_i$, where $s_i$ denotes structured signal and $n_i$ zero-mean perturbations. A shape-preserving reduction operator $P$ maps the window to a reduced representation (e.g., via merging or downsampling). Under a local linearization of the block, and for operators such as IE-KVD that are explicitly linear in each neighborhood, we approximate its effect via a \emph{windowed frequency response} $H_P(k)$:
\begin{equation}
\begin{aligned}
\mu'
&\approx
\sum_k w_k H_P(k)\,\mu_k,\\
\sigma'^2
&\approx
\sum_k w_k^2 H_P(k)^2\,\sigma_k^2.
\end{aligned}
\label{eq:mu-sigma-prime}
\end{equation}

This spectral model provides a useful approximation of how token-reduction operators reshape the band-weighted paired statistic. For strictly linear operators such as IE-KVD the frequency response $H_P(k)$ is exact, while for merging-based operators (e.g., L-GTM, ABM), it serves as a local linearization that captures their dominant low-pass-like behaviour. 

In particular, any operator $P$ whose effective response $H_P(k)$ (exact for IE-KVD, approximate for merging) acts as a \emph{bandwise shrinkage rule}, with $H_P(k)\approx 1$ on margin-rich bands and $H_P(k)<1$ on variance-heavy bands, will tend to reduce the ratio $r=\sigma/\mu$ in \eqref{eq:cantelli}. Both of our proposed modules, Laplacian-gated token merging (L-GTM) and Interpolate--Extrapolate KV-Downsampling (IE-KVD), are designed to approximate this behavior: they selectively smooth or merge tokens that are redundant while retaining tokens that appear informative, producing a frequency-selective shrinkage profile.

For IE-KVD, which is a linear filtered downsampling operator, the spectral interpretation is exact, for merging-based operators, it should be viewed as an approximation under local linearization. We emphasize that this analysis is a stylized model: it relies on linearization and approximate bandwise decorrelation, and is intended to clarify the design principle behind our frequency-aware token reduction rather than to provide a formal guarantee.

\subsection{A Margin--variance Improvement Criterion}

We define the changes in mean margin and variance as
\begin{equation}
\begin{aligned}
\Delta\mu
&:=
\mu - \mu',\\
\Delta\sigma^2
&:=
\sigma^2 - \sigma'^2.
\end{aligned}
\label{eq:deltas}
\end{equation}
We are interested in when the ratio $r'=\sigma'/\mu'$ is smaller than $r=\sigma/\mu$, since by \eqref{eq:cantelli} this implies a tighter Cantelli bound.

\begin{theorem}[Spectral margin--variance improvement]
\label{thm:spectral-improvement}
Assume $\mu>0$, $\mu'>0$, $\sigma^2>0$, and that $\mu'$ and $\sigma'^2$ are defined as above. Then the post-reduction ratio $r'$ is smaller than the original ratio $r$, \( r' < r, \) if and only if
\begin{equation}
\Delta\sigma^2
\;>\;
2\,\frac{\sigma^2}{\mu}\,\Delta\mu
\;-\;
\frac{\sigma^2}{\mu^2}\,(\Delta\mu)^2.
\label{eq:exact-condition}
\end{equation}
Moreover, when $|\Delta\mu| \ll \mu$, the first-order sufficient condition
\begin{equation}
\Delta\sigma^2
\;>\;
2\,\frac{\sigma^2}{\mu}\,\Delta\mu
\label{eq:first-order-condition}
\end{equation}
guarantees $r'<r$ and hence a strictly improved Cantelli bound.
\end{theorem}

\paragraph{Proof.}
Since $\mu>0$ and $\mu'>0$ we have $r,r'\ge 0$, so $r'<r$ is equivalent to $r'^2<r^2$. The condition $r'^2<r^2$ is equivalent to
\(
\frac{\sigma^2-\Delta\sigma^2}{(\mu-\Delta\mu)^2}
<
\frac{\sigma^2}{\mu^2}.
\)
Multiplying both sides by the positive denominators and
rearranging gives
\(
\Delta\sigma^2\mu^2 - 2\sigma^2\mu\,\Delta\mu
+ \sigma^2(\Delta\mu)^2 > 0,
\)
which after dividing by $\mu^2$ yields
\eqref{eq:exact-condition}.
Expanding the right-hand side of
\eqref{eq:exact-condition} to first order in $\Delta\mu/\mu$
gives the sufficient condition \eqref{eq:first-order-condition}.
\hfill$\square$

\subsection{Interpretation and Scope}

\autoref{eq:exact-condition}, together with the bandwise decompositions of $\Delta\mu$ and $\Delta\sigma^2$, provides a spectral lens for comparing reduction strategies. The term $\Delta\sigma^{2} = \sum_{k} w_{k}^{2} (1 - H_{P}(k)^{2})\,\sigma_{k}^{2}$ captures the \emph{variance savings}, while $\Delta\mu = \sum_{k} w_{k} (1 - H_{P}(k))\,\mu_{k}$ quantifies the corresponding \emph{margin cost}. Achieving a tighter decision bound ($r' < r$) requires maximizing the former while keeping the latter small.

Standard token-merging approaches (e.g., ToMe~\cite{bolya2023tokenmergingfaststable}) can be interpreted, in our stylized spectral model, as performing spatial averaging: they tend to smooth fine-scale structure and thus resemble a low-pass filter with an effective response $H_{P}(k) \ll 1$ for high-frequency bands $k$. Because edges, textures, and other fine-scale structures often yield large $\mu_{k}$ (i.e., high-frequency bands are \emph{margin-rich} for recognition), indiscriminate merging induces a substantial margin loss $\Delta\mu$. If this loss exceeds the allowable threshold in Theorem~\ref{thm:spectral-improvement}, the classifier’s performance degrades despite any reduction in variance.

In contrast, the proposed {\bf BiGain} is designed to favor this trade-off. Laplacian gating identifies regions where high-frequency content is relatively small, which in our spectral model corresponds to bands with small $\mu_{k}$. In regions containing significant high-frequency structure (large $\mu_{k}$), {\bf BiGain} suppresses merging, corresponding to an effective response $H_{P}(k) \approx 1$ and hence $\Delta\mu \approx 0$ on margin-critical bands in our spectral model. This frequency-aware selection biases the shrinkage profile towards satisfying \eqref{eq:exact-condition}, which is consistent with our observation that \textsc{BiGain} preserves discriminative utility far better than spectrally agnostic merging schemes.

\section{Implementation Details}

\subsection{Datasets and Evaluation Protocols}

\subsubsection{Dataset Details}
\label{app:dataset}

We evaluate on four widely-used benchmarks, summarized in Table~\ref{tab:dataset_stats}. Following \cite{li2023your}, ImageNet-1K is sub-sampled to 2,000 images for classification to reduce computational cost, while the full validation set is retained for generation experiments.

\begin{table*}[!htb]
\centering
\caption{Dataset statistics with official splits used in our experiments.}
\label{tab:dataset_stats}
\vspace{-0.08in}
\resizebox{0.53\textwidth}{!}{
\begin{tabular}{lcccc}
\toprule
Dataset & Classes & Split & \# Images (Cls.) & \# Images (Gen.) \\
\midrule
ImageNet-100 \cite{tian2020contrastive} & 100 & Val. & 5{,}000 & 5{,}000 \\
ImageNet-1K \cite{russakovsky2015imagenet} & 1,000 & Val. & 2{,}000 & 50{,}000 \\
Oxford-IIIT Pets \cite{parkhi2012cats} & 37 & Test & 3{,}669 & 3{,}669 \\
COCO-2017 \cite{lin2014microsoft} & 80 & Val. & 5{,}000 & 5{,}000 \\
\bottomrule
\end{tabular}
}
\end{table*}

\subsubsection{Diffusion Classifier Protocol}
\label{app:adc_params}

\paragraph{Diffusion-classifier.}
We follow the \emph{Diffusion Classifier} framework~\cite{li2023your}, which scores a candidate conditioning $c$ by the expected noise-prediction error $\E_{t,\epsilon}\!\left[\lVert \epsilon - \epsilon_\theta(x_t,c,t)\rVert_2^2\right]$ and selects the minimizer. This method is \emph{training-free}, requiring no calibration or finetuning, and enables zero-shot classification directly from pretrained diffusion models. To enable evaluation on large label spaces, we use adaptive evaluation with staged pruning (detailed in Algorithm~\ref{app:adc_alg}). We adjust only \texttt{TrialList} and \texttt{KeepList} based on the size of the candidate set.

\begin{table*}[!htb]
\centering
\caption{Adaptive diffusion-classifier parameters per dataset. $N_{\text{stages}}$ is the number of pruning stages; \texttt{TrialList} is the cumulative number of Monte Carlo trials per candidate by stage; \texttt{KeepList} is the number of candidates retained after each stage.}
\label{tab:adc_params}
\vspace{-0.08in}
\resizebox{0.42\textwidth}{!}{
\begin{tabular}{lccc}
\toprule
Dataset & $N_{\text{stages}}$ & \texttt{TrialList} & \texttt{KeepList} \\
\midrule
ImageNet-100 & 2 & [5, 20] & [5, 1] \\
COCO-2017 & 2 & [5, 20] & [5, 1] \\
Oxford-IIIT Pets & 2 & [5, 20] & [5, 1] \\
ImageNet-1K & 3 & [5, 20, 100] & [50, 10, 1] \\
\bottomrule
\end{tabular}
}
\end{table*}

For completeness, we also evaluated velocity-prediction flow-matching models (FLUX~\cite{labs2025flux1kontextflowmatching}). Using the \texttt{FlowMatchEulerDiscreteScheduler} to construct affine mappings for recovering $\hat\epsilon_\theta$ and $\hat x_0$ within DDIM, the released FLUX.1-dev checkpoint performed only marginally better than random guessing under the diffusion-classifier protocol. To avoid adapter-specific confounds and ensure a fair comparison, we restrict all evaluations to standard noise-prediction models.

\subsection{Multi-Label Classification Metric}
\label{app:multilabel_map}

We evaluate performance using mean Average Precision (mAP) for multi-label image classification,
computed from a ranked list of labels for each image.

Given an image $x$ with ground-truth label set $\mathcal{Y} \subset \mathcal{C}$, the diffusion classifier assigns each label $c \in \mathcal{C}$ an estimated classification error
\begin{equation}
\hat L(x,c) = \frac{1}{S} \sum_{s=1}^{S} \ell(x,c; t_s, \epsilon_s),
\end{equation}
where $\ell(x,c;t_s,\epsilon_s)$ and the shared Monte Carlo set $\{(t_s,\epsilon_s)\}_{s=1}^S$ are defined in Sec.~3.1.1 of the main paper. Lower values of $\hat L(x,c)$ indicate higher confidence that label $c$ is present.

All labels are ranked in ascending order of $\hat L(x,c)$.

\paragraph{Average Precision per image.}
Let $\{c_1, c_2, \dots\}$ denote the ranked label list. The Average Precision for image $x$ is defined as
\begin{equation}
\mathrm{AP}(x)
= \frac{1}{|\mathcal{Y}|}
\sum_{k:\, c_k \in \mathcal{Y}}
\frac{|\mathcal{Y} \cap \{c_1,\dots,c_k\}|}{k},
\end{equation}
where the sum runs over ranks at which a ground-truth label appears.

\paragraph{Mean Average Precision.}
The final mAP is obtained by averaging $\mathrm{AP}(x)$ over all test images.

\subsection{Model Configurations}

\subsubsection{Prompt Templates}

For the classification task, following~\cite{li2023your}, we use \texttt{``a photo of a \{class\}''} for ImageNet and COCO datasets, and \texttt{``a photo of a \{class\}, a type of pet''} for Oxford-IIIT Pets.

For generation, we use the same templates except for COCO-2017, where we use the official validation captions.

\subsubsection{Generation Setup}
\label{app:generation}

We standardize generation across both backbones. For Stable Diffusion 2.0 (UNet)~\cite{rombach2022high}, we use the EulerDiscreteScheduler with a scaled-linear beta schedule (beta\_start 0.00085, beta\_end 0.012, 1{,}000 training steps, epsilon prediction). For DiT-XL/2-512~\cite{peebles2023scalable}, we use the DDIMScheduler with a linear beta schedule (beta\_start 0.0001, beta\_end 0.02, 1{,}000 training steps, epsilon prediction). In both cases, we sample for 50 steps at 512×512 resolution. We apply classifier-free guidance with a scale of 7.5 for Stable Diffusion 2.0 and 4.0 for DiT-XL/2-512. Unless otherwise stated, all experiments are conducted in FP16 precision. For evaluation, FID scores are computed using the \texttt{pytorch-fid} implementation~\cite{Seitzer2020FID}.

\subsection{Token Compression}
\label{app:token_compress_setup}

\subsubsection{Compression Settings}
\label{app:token_default}

Guided by the ablation in Table~\ref{tab:pets-ablation-modules}, we apply compression exclusively to self-attention (SA) and leave cross-attention (CA) and MLP blocks intact to preserve prompt adherence.  For merging-based operators, merging is performed inside each SA block and an explicit unmerge restores the original sequence length before the residual addition, ensuring dense outputs for downstream modules. For KV-downsampling operators, only keys and values are subsampled while queries remain full-length, removing the need for unmerge. 

\noindent\textbf{Stable Diffusion~2.0 (U\!-Net).}
We insert compression exclusively at the highest-resolution encoder layers, where the spatial token count, and thus attention cost is maximal. 
This targets the primary bottleneck while maintaining quality.

\noindent\textbf{Diffusion Transformer (DiT-XL/2).}
To assess generality beyond U-Net architectures, we port the same operators to DiT-XL/2. Specifically, token compression is applied within the first 12 transformer blocks, comparing early (blocks 1--6) versus mid-early (blocks 7--12) reduction, while leaving later blocks, where class conditioning and fine structural details consolidate unchanged.

\subsubsection{Baseline Implementation}
\label{app:baselines}

For all token compression baselines, we use the official implementations and default parameters released by the authors, and run them under a common experimental protocol (see Sec.~\ref{app:adc_params} and Sec.~\ref{app:generation}) to ensure fair comparison and avoid unintentional re-tuning. The only modification we introduce is to vary the token reduction ratio, so that each method can be fairly evaluated under different levels of compression.

\subsection{Efficiency Evaluation}
\label{app:efficiency}

To measure the acceleration effect of our token reduction methods, we evaluate on the official Stable Diffusion~2.0 implementation released by Stability AI \cite{rombach2022high}. All experiments are conducted on a single NVIDIA RTX~4090 GPU in half-precision (\texttt{float16}). We report wall-clock sampling time per image batch excluding the VAE encoding/decoding overhead, since our methods target the denoising backbone rather than the autoencoder. \textsc{FLOPs} are measured using \texttt{FlopCounterMode} from \texttt{torch.utils.flop\_counter}~\cite{paszke2019pytorchimperativestylehighperformance}. The corresponding runtime and efficiency results are summarized in Table \ref{tab:sd20-efficiency-bs4}, which demonstrates the advantage of our method.

\begin{table*}[!h]
\centering
\small
\caption{\textbf{Stable Diffusion~2.0 efficiency (batch size 4).} Wall-clock sampling time per \emph{batch} (seconds) excluding VAE encode/decode. All rows use merge ratio $r=0.7$.}
\label{tab:sd20-efficiency-bs4}
\vspace{-0.08in}
\begin{tabular}{lcccc}
\toprule
\textbf{Method} & \textbf{Time $\downarrow$ (s / batch)} & \textbf{Acceleration $\uparrow$ (\text{\%})} & \textbf{FLOPs $\downarrow$ (G)} \\
\midrule
Baseline (No Accel.)        & 11.98 & --    & 804.26 \\
SiTo~\cite{zhang2025training}    &  8.71 & 27.30 & 748.49 \\
ToMe~\cite{bolya2023tokenmergingfaststable}        &  7.37 & 38.48 & 704.87 \\
Laplacian Gated Merge (Ours)       &  7.37 & 38.48 & 704.99 \\
Cached Assignment Merge (Ours)    &  7.29 & 39.15 & 698.88 \\
Adaptive Block Merging (Ours)     &  7.27 & 39.32 & 695.08 \\
\bottomrule
\end{tabular}
\end{table*}

\section{Algorithm}
\label{app:adc}

\subsection{Adaptive Diffusion Classifier}
\label{app:adc_alg}

Naïve diffusion classification requires evaluating all candidate classes, and thus its cost grows linearly with the number of classes. To mitigate this, we adopt the adaptive evaluation strategy introduced in the diffusion-classifier framework \cite{li2023your}. At each stage, we allocate a fixed budget of trials across the remaining classes, discard unlikely candidates based on their average error, and retain only the most promising ones. This progressive pruning concentrates computation on high-confidence classes, enabling more fine-grained Monte Carlo error estimation. The procedure is summarized in Algorithm~\ref{alg:adaptive_diffusion_classifier}.

\begin{algorithm}[!h]
\caption{Diffusion Classifier (Adaptive) \cite{li2023your} }
\label{alg:adaptive_diffusion_classifier}
\begin{algorithmic}[1]
\Require test image $\mathbf{x}$, conditioning inputs $\mathcal{C} = \{\mathbf{c}_i\}_{i=1}^n$ (e.g., text embeddings or class indices), number of stages $N_{\text{stages}}$, list \texttt{KeepList} of number of $\mathbf{c}_i$ to keep after each stage, list \texttt{TrialList} of number of trials done by each stage
\State Initialize $\texttt{Errors}[\mathbf{c}_i] = \text{list}()$ for each $\mathbf{c}_i$
\State Initialize $\texttt{PrevTrials} = 0$ \Comment{How many times we've tried each remaining element of $\mathcal{C}$ so far}
\For{stage $i = 1, \ldots, N_{\text{stages}}$}
    \For{trial $j = 1, \ldots, \texttt{TrialList}[i] - \texttt{PrevTrials}$}
        \State Sample $t \sim [1, 1000]$
        \State Sample $\boldsymbol{\epsilon} \sim \mathcal{N}(0, I)$
        \State $\mathbf{x}_t = \sqrt{\bar{\alpha}_t}\mathbf{x} + \sqrt{1 - \bar{\alpha}_t}\boldsymbol{\epsilon}$
        \For{conditioning $\mathbf{c}_k \in \mathcal{C}$}
            \State $\texttt{Errors}[\mathbf{c}_k].\text{append}(\|\boldsymbol{\epsilon} - \boldsymbol{\epsilon}_\theta(\mathbf{x}_t, \mathbf{c}_k, t)\|^2)$
        \EndFor
    \EndFor
    \State $\mathcal{C} \leftarrow \underset{\substack{\mathcal{S} \subset \mathcal{C}; \\ |\mathcal{S}| = \texttt{KeepList}[i]}}{\arg\min} \sum_{\mathbf{c}_k \in \mathcal{S}} \text{mean}(\texttt{Errors}[\mathbf{c}_k])$ \Comment{Keep top $\texttt{KeepList}[i]$ conditionings}
    \State $\texttt{PrevTrials} = \texttt{TrialList}[i]$
\EndFor
\State \Return $\underset{\mathbf{c}_i \in \mathcal{C}}{\arg\min} \text{mean}(\texttt{Errors}[\mathbf{c}_i])$
\end{algorithmic}
\end{algorithm}

\subsection{Frequency-Aware Token Scoring} 
\label{app:freq_scores}

\emph{Spectral} structure of latent features is important for both discriminative and generative ability. High-frequency tokens encode the information of edges, textures, and small objects, especially at the late denoise stage, which are indispensable for recognition. However, high-frequency tokens can also amplify the variance in the diffusion classifier since predictions are aggregated over Monte Carlo draws of timesteps and noise; excess high-frequency tokens inflate the per-timestep estimation variance. Moreover, different timesteps emphasize different bands, early denoising focuses on low frequencies (global structure) while later steps emphasize high frequencies (fine detail). Therefore, the compression schedule should be \emph{spectrally balanced and temporally consistent} to avoid injecting avoidable variance across timesteps. The necessity of preserving a balanced spectrum is confirmed empirically in Table~\ref{append:imagenet100_freq_down}, where discarding either high- or low-frequency tokens severely harms classification.

\begin{table*}[!ht]
\centering
\small
\label{app:frequency-downsampling}
\caption{\textbf{Classification results on frequency-based KV selection on ImageNet-100.} 
We compare the standard \textsc{ToDo} strategy with frequency-aware variants that select tokens with the highest or lowest Laplacian scores globally. Retaining only high- or low-frequency tokens severely degrades classification performance, highlighting the need to preserve a balanced spectrum.}
\label{append:imagenet100_freq_down}
\begin{tabular}{l cc}
\toprule
\textbf{Downsampling strategy} & \textbf{Acc@1 $\uparrow$} & KV token sparsity\\
\midrule
Todo (Nearest-Neighbor)~\cite{smith2024todo} & 72.30 & 75\% \\
Low-frequency tokens (lowest-laplacian) & 45.58 & 75\% \\
High-frequency tokens (Highest-laplacian) & 26.56 & 75\% \\
\bottomrule
\end{tabular}
\end{table*}

Our \textbf{BiGain$_{\text{TM}}$} design follows this principle. Since token merging resembles a local low-pass filter, we encourage merging only in small, spectrally smooth neighborhoods, where low-frequency information can be safely aggregated, while protecting detail-rich tokens that anchor class-critical microstructures. This balanced policy removes redundancy without sacrificing classification accuracy or generation fidelity. Practically, we introduce a set of fast, training-free scoring heuristics to decide which tokens to \emph{preserve} (high detail) and which to \emph{merge} (smooth/redundant), and we apply them consistently across timesteps so that each per-timestep classifier score remains reliable and contributes coherently to the Monte Carlo ensemble.

\paragraph{Notation.}
Let $\mX \in \R^{H\times W\times C}$ denote the hidden feature tensor (height $H$, width $W$, channels $C$). For spatial index $(i,j)$, the token (channel vector) is $\vx_{i,j}\!:=\mX_{i,j,:}\in\R^C$. The global mean token is $\vmu := \tfrac{1}{HW}\sum_{p=1}^{H}\sum_{q=1}^{W}\vx_{p,q}$. For a $3{\times}3$ spatial kernel $\mL$, $(\mX \ast \mL)_{i,j,c}$ denotes 2-D convolution at $(i,j)$ on channel $c$. Let $\sN_4(i,j)$ be the (in-bounds) 4-neighborhood of $(i,j)$ (up/down/left/right). The DFT of $\vx_{i,j}$ at channel-frequency bin $k$ is $\hat{\vx}_{i,j,k} := \sum_{c=1}^{C}(\vx_{i,j})_c\,e^{-2\pi \mathrm{i}\, (c-1)(k-1)/C}$ for $k \in \{1,\ldots,C\}$. We write $||\cdot||_p$ for the vector $\ell_p$ norm, $||\cdot||\equiv||\cdot||_2$, and $\langle\va,\vb\rangle$ for the Euclidean inner product. We compute a scalar score $F_{i,j}\in\R$ per token, where larger values indicate detail-rich tokens and smaller values indicate smooth/redundant tokens. We list all functions of different metrics in Table ~\ref{append:features}.

\begin{table*}[htbp]
\centering
\caption{Formulas of different metrics.}
\label{append:features}
\begin{tabular}{@{}ll@{}}
\toprule
\textbf{Metric Name} & \textbf{Formula} \\
\midrule
Global mean deviation & $F_{i,j} = \| \boldsymbol{x}_{i,j} - \boldsymbol{\mu} \|$ \\
$\ell_1$ norm & $F_{i,j} = \| \boldsymbol{x}_{i,j} \|_1$ \\
$\ell_2$ norm & $F_{i,j} = \| \boldsymbol{x}_{i,j} \|$ \\
Channel variance & $F_{i,j} = \frac{1}{C} \sum_{c=1}^{C} \left( (\boldsymbol{x}_{i,j})_c - \frac{1}{C} \sum_{c'=1}^{C} (\boldsymbol{x}_{i,j})_{c'} \right)^2$ \\
Laplacian ($\ell_1$) & $F_{i,j} = \frac{1}{C} \sum_{c=1}^{C} \left| (\mX * \mL)_{i,j,c} \right|$, \quad $\mL = \begin{bmatrix} 0 & 1 & 0 \\ 1 & -4 & 1 \\ 0 & 1 & 0 \end{bmatrix}$ \\
Laplacian ($\ell_2$) & $F_{i,j} = \sqrt{ \frac{1}{C} \sum_{c=1}^{C} \left( (\mX * \mL)_{i,j,c} \right)^2 }$ \\
DFT spectral centroid & $F_{i,j} = \frac{ \sum_{k=1}^{C} k \, | \hat{\boldsymbol{x}}_{i,j,k} | }{ \sum_{k=1}^{C} | \hat{\boldsymbol{x}}_{i,j,k} | }$ \\
DFT total amplitude & $F_{i,j} = \sum_{k=1}^{C} | \hat{\boldsymbol{x}}_{i,j,k} |$ \\
Cosine similarity to neighbors & $F_{i,j} = \frac{1}{|\mathbb{N}_4(i,j)|} \sum_{(p,q) \in \mathbb{N}_4(i,j)} \frac{ \langle \boldsymbol{x}_{i,j}, \boldsymbol{x}_{p,q} \rangle }{ \| \boldsymbol{x}_{i,j} \| \, \| \boldsymbol{x}_{p,q} \| }$ \\
Cosine similarity to global mean & $F_{i,j} = \frac{ \langle \boldsymbol{x}_{i,j}, \boldsymbol{\mu} \rangle }{ \| \boldsymbol{x}_{i,j} \| \, \| \boldsymbol{\mu} \| }$ \\
\bottomrule
\end{tabular}
\end{table*}

For all heuristics except cosine-based ones, larger $F_{i,j}$ indicates stronger local variation and thus high-frequency detail. In contrast, for cosine similarity scores, \emph{smaller} values correspond to tokens that deviate more from their neighbors or the global mean, and are therefore detail-rich.

\subsection{\texorpdfstring{BiGain$_{\textbf{TM}}$}{BiGain TM}}
\label{app:bigaintm_alg}

Algorithm~\ref{alg:bigain_tm} presents our frequency-aware token merging method. The core innovation lies in using spectral information to guide merge decisions, ensuring that token reduction preserves both generative fidelity and discriminative utility. The algorithm first applies a frequency scorer $\mathcal{F}$ (default: Laplacian filtering~\ref{app:freq_scores}) to identify local frequency content in the spatial feature map. Tokens with low frequency scores indicate smooth, homogeneous regions amenable to merging, while high scores correspond to edges, textures, and fine details critical for classification.

The destination selection step partitions the spatial layout into regular grids and identifies the lowest-frequency token within each grid as a merge destination. This strategy ensures spatial coverage while directing merging toward spectrally smooth regions. The remaining tokens form a source set, which is then assigned to destinations via bipartite matching based on cosine similarity. By selecting the top-$r$ fraction of most similar pairs, the method preserves semantic coherence while respecting the frequency-based partitioning. After merging and processing through attention layers, an unmerge operation restores the original sequence length for architectural compatibility.

\begin{algorithm}[!htbp]
\caption{{\bf BiGain$_\texttt{TM}$}: Frequency-Aware Token Merging}
\label{alg:bigain_tm}
\begin{algorithmic}[1]
\Require Tokens $\mX \in \R^{N \times d}$, merge ratio $r$, grid size $s$, frequency scorer $\mathcal{F}$
\Function{BiGainMerge}{$\mX$, $r$, $s$, $\mathcal{F}$}
\State $\vf \leftarrow \mathcal{F}(\mX)$ \Comment{Score tokens by frequency content}
\State $\sD \leftarrow \text{SelectDestinations}(\vf, s)$ \Comment{Lowest frequency per grid}
\State $\sS \leftarrow \{1, \ldots, N\} \setminus \sD$ \Comment{Remaining tokens as sources}
\State $\mathcal{M} \leftarrow \text{BipartiteMatch}(\mX_\sS, \mX_\sD, r)$ \Comment{Similarity-based assignment}
\State $\mX^{\text{merged}} \leftarrow \text{Merge}(\mX, \mathcal{M})$ \Comment{Combine assigned tokens}
\State $\mZ \leftarrow \text{Process}(\mX^{\text{merged}})$ \Comment{Apply attention}
\State \Return $\text{Unmerge}(\mZ, \mathcal{M})$ \Comment{Restore dimensions}
\EndFunction
\end{algorithmic}
\end{algorithm}

Algorithm~\ref{alg:adaptive_block_merge} presents Adaptive Block Merge (ABM), a computationally efficient variant designed for high-resolution stages where token count is maximal. Rather than per-token assignment, ABM operates at block granularity. After computing frequency scores, the feature map is partitioned into blocks, and blocks are ranked by their frequency content. The lowest-scoring fraction $r$ of blocks are identified as smooth regions and merged via averaging, while high-frequency blocks remain intact. This block-level decision reduces computational complexity of bipartite matching, providing speedup with little accuracy degradation as demonstrated in our Table~\ref{tab:sd20-merge_classification_ana}.

\begin{algorithm}[!htbp]
\caption{Adaptive Block Merge (ABM): Fast {\bf BiGain$_\texttt{TM}$} Variant}
\label{alg:adaptive_block_merge}
\begin{algorithmic}[1]
\Require Tokens $\mX \in \R^{N \times d}$, block size $b$, merge ratio $r \in [0,1]$, scorer $\mathcal{F}$
\Function{AdaptiveBlockMerge}{$\mX$, $b$, $r$, $\mathcal{F}$}
\State $\vf \leftarrow \mathcal{F}(\mX)$ \Comment{Compute frequency scores}
\State $\mathcal{B} \leftarrow \text{BlockPartition}(\mX, b)$ \Comment{Partition into $b \times b$ blocks}
\State $\mathcal{B}_{\text{smooth}} \leftarrow \text{SelectLowestFreq}(\mathcal{B}, \vf, r)$ \Comment{Select lowest $r$ fraction blocks}
\State $\mX^{\text{merged}} \leftarrow \text{MergeBlocks}(\mX, \mathcal{B}_{\text{smooth}})$ \Comment{Average selected blocks}
\State $\mZ \leftarrow \text{Process}(\mX^{\text{merged}})$ \Comment{Apply attention}
\State \Return $\text{RestoreBlocks}(\mZ, \mathcal{B}_{\text{smooth}})$ \Comment{Restore dimensions}
\EndFunction
\end{algorithmic}
\end{algorithm}

\subsection{\texorpdfstring{BiGain$_{\textbf{TD}}$}{BiGain TD}}
\label{app:bigaintd_alg}

Algorithm~\ref{alg:bigain_td} presents our Interpolate-Extrapolate KV-Downsampling method, which reduces attention complexity by downsampling keys and values while preserving queries at full resolution. This asymmetric approach maintains the model's ability to attend precisely to all spatial positions while reducing memory and computation. The key innovation is the controllable linear combination of nearest-neighbor and average pooling, allowing fine-grained control over the frequency-preservation trade-off.

Here we use the same interpolate-extrapolate operator $D_{\alpha,s}$ as defined in Eq.~\ref{eq:kv_operator} of the main paper. This operator blends nearest-neighbor sampling (preserving detail) with average pooling (smoothing), controlled by the parameter $\alpha \in \mathbb{R}$. Keys and values are downsampled as $\tilde{K} = D_{\alpha,s}(K)$ and $\tilde{V} = D_{\alpha,s}(V)$, while queries remain full resolution.

\begin{algorithm}[!h]
\caption{{\bf BiGain$_\texttt{TD}$}: Interpolate-Extrapolate KV-Downsampling (IE-KVD)}
\label{alg:bigain_td}
\begin{algorithmic}[1]
\Require Tokens $\mX \in \R^{N \times d}$, downsample factor $s$, interpolation-extrapolation factor $\alpha \in \mathbb{R}$
\Function{BiGainDownsample}{$\mX$, $s$, $\alpha$}
\State $\mQ \leftarrow \mX \mW_Q$ \Comment{Queries at full resolution}
\State $\mK \leftarrow \mX \mW_K,\quad \mV \leftarrow \mX \mW_V$ \Comment{Keys and values}
\State $\tilde{\mK} \leftarrow \text{Interpolate/ExtrapolateDownsample}(\mK, s, \alpha)$
\State $\tilde{\mV} \leftarrow \text{Interpolate/ExtrapolateDownsample}(\mV, s, \alpha)$
\State $\mZ \leftarrow \text{Attention}(\mQ, \tilde{\mK}, \tilde{\mV})$
\State \Return $\mZ$ \Comment{Output remains at full resolution}
\EndFunction
\end{algorithmic}
\end{algorithm}

\begin{table*}[t]
\centering
\small
\caption{\textbf{ImageNet-1K subset robustness.}
Diffusion-classifier accuracy on 2K and 10K subsets with 95\% Wilson confidence intervals.}
\label{tab:imagenet_subset_ci}
\begin{tabular}{lccccc}
\toprule
Method & No accel. & ToMe (70\%) & BiGain$_{\text{TM}}$ (70\%) & ToDo (2$\times$) & BiGain$_{\text{TD}}$ (2$\times$) \\
\midrule
Acc. ($n{=}2000$) (\%) $\uparrow$ & $57.05\pm2.17$ & $37.35\pm2.13$ & $44.50\pm2.19$ & $55.75\pm2.17$ & $56.30\pm2.17$ \\
Acc. ($n{=}10000$) (\%) $\uparrow$ & $57.92\pm0.97$ & $39.85\pm0.96$ & $45.25\pm0.98$ & $56.50\pm0.97$ & $57.59\pm0.97$ \\
\bottomrule
\end{tabular}
\end{table*}

\begin{table*}[t]
\centering
\small
\caption{\textbf{IE-KVD $\alpha$ sensitivity.}
SD-2.0 / Oxford-IIIT Pets diffusion classification accuracy (Acc., \%).}
\label{tab:iekvd_alpha}
\begin{tabular}{lcccccc}
\toprule
$\alpha$ & 0.0 & 0.5 & 0.8 & 0.9 ({\bf BiGain$_{\texttt{TD}}$}) & 1.0 & 1.2 \\
\midrule
Acc. ($s{=}2$) (\%) $\uparrow$ & 77.02 & 78.74 & 79.97 & 81.52 & 81.30 & 64.27 \\
Acc. ($s{=}4$) (\%) $\uparrow$ & 71.26 & 75.63 & 77.48 & 78.03 & 77.46 & 45.46 \\
\bottomrule
\end{tabular}
\end{table*}

\begin{table}[t]
\centering
\small
\caption{\textbf{IE-KVD schedule ablation.}
SD-2.0 / Oxford-IIIT Pets generation quality (FID $\downarrow$) under different linear schedules.}
\label{tab:iekvd_schedule}
\begin{tabular}{lc}
\toprule
Linear schedule $(\alpha_{\text{start}}\!\rightarrow\!\alpha_{\text{end}})$ & FID $\downarrow$ \\
\midrule
ToDo & 33.52 \\
\midrule
IE-KVD: $0.2\!\rightarrow\!0.8$ & 35.56 \\
IE-KVD: $0.5\!\rightarrow\!1.0$ & 33.95 \\
IE-KVD: $0.7\!\rightarrow\!1.0$ & 33.89 \\
IE-KVD: $0.8\!\rightarrow\!1.2$ & 32.19 \\
IE-KVD: $0.0\!\rightarrow\!1.2$ & 32.54 \\
\bottomrule
\end{tabular}
\end{table}

\subsection{Additional Ablations and Results}
\label{sec:add_ablation_results}

We provide four targeted supplementary studies to further clarify the robustness and scope of our results: (i) robustness to the ImageNet-1K evaluation subset size, (ii) sensitivity of IE-KVD to $\alpha$ and its timestep schedule, (iii) compatibility between the merging and downsampling modules, and (iv) actual wall-clock speedups. Throughout this subsection, {\bf BiGain$_{\text{TM}}$} refers to the merging module (L-GTM), while {\bf BiGain$_{\text{TD}}$} refers to the KV-downsampling module (IE-KVD). Unless otherwise stated, we follow the same evaluation protocol as in the main paper. For SD-2.0, we report diffusion-classifier Top-1 accuracy (Acc@1, \%) and generation quality (FID). For merging-based methods, the merge ratio is denoted by $r$; for downsampling-based methods, the downsampling factor is denoted by $s$. For IE-KVD, $\alpha$ controls interpolation/extrapolation in KV downsampling; for generation we also consider linear schedules $(\alpha_{\text{start}}\!\rightarrow\!\alpha_{\text{end}})$. Wall-clock times are measured on a single RTX~4090 with FP16 and batch size 4 over 50 denoising steps, we report the U-Net step time averaged over 50 runs (VAE excluded).

\paragraph{Robustness to evaluation subset size.}
Table~\ref{tab:imagenet_subset_ci} addresses the concern that the ImageNet-1K classification results in the main paper are evaluated on a 2K subset for efficiency. We therefore expand the evaluation to 10K images in Table~\ref{tab:imagenet_subset_ci} and find that the relative ranking of methods remains unchanged, leading to the same qualitative conclusion. This indicates that the main findings are robust and not an artifact of the smaller subset.

\paragraph{Sensitivity of IE-KVD hyperparameters.}
Tables~\ref{tab:iekvd_alpha} and~\ref{tab:iekvd_schedule} examine the two main IE-KVD design choices: the interpolation--extrapolation parameter $\alpha$ and its timestep schedule. Accuracy is strongest around $\alpha \in [0.8,1.0]$, showing that the default choice $\alpha=0.9$ is not fragile, while overly aggressive extrapolation ($\alpha=1.2$) degrades performance. For generation, FID varies only modestly across linear schedules, suggesting that IE-KVD is reasonably robust to the exact schedule.

\noindent{\bf Joint use of merging and downsampling.}
Table~\ref{tab:joint_perf} studies two hybrid placements: L-GTM in the encoder with IE-KVD in the decoder, and the reverse. Both combinations remain stable for classification and generation, but neither exceeds the best single-module setting. This suggests that the two operators are compatible, although their gains are not simply additive.

\noindent{\bf Measured runtime.}
Finally, Table~\ref{tab:wallclock_ms} reports actual U-Net step time in addition to FLOPs. This complements Table~\ref{tab:sd20-efficiency-bs4}, which focuses on merging-based batch-time measurements, by providing a unified runtime comparison that also includes downsampling and hybrid settings. The empirical speedups follow the same trend as the FLOP reductions, confirming that the proposed operators translate into real inference-time gains.

\begin{table}[!ht]
\centering
\small
\caption{\textbf{Joint use of L-GTM and IE-KVD.}
SD-2.0 / Oxford-IIIT Pets with $r{=}0.7$, $s{=}2$, $\alpha{=}0.9$.}
\label{tab:joint_perf}
\begin{tabular}{lcc}
\toprule
Method & Acc@1 $\uparrow$ (\%) & FID $\downarrow$ \\
\midrule
No accel. & 81.03 & 35.01 \\
ToMe & 65.76 & 38.35 \\
ToDo & 81.30 & 33.52 \\
BiGain$_{\text{TM}}$ & 74.63 & 37.73 \\
BiGain$_{\text{TD}}$ & 81.52 & 32.19 \\
\midrule
L-GTM(enc) + IE-KVD(dec) & 79.53 & 34.84 \\
IE-KVD(enc) + L-GTM(dec) & 79.23 & 36.90 \\
\bottomrule
\end{tabular}
\end{table}

\begin{table}[!ht]
\centering
\small
\caption{\textbf{Wall-clock UNet step time.}
SD-2.0 on RTX 4090 (FP16, batch{=}4, 50 steps; averaged over 50 runs; VAE excluded).}
\label{tab:wallclock_ms}
\resizebox{\linewidth}{!}{
\begin{tabular}{lccc}
\toprule
Method & Time $\downarrow$ (ms/step) & Speedup $\uparrow$ ($\times$) & FLOPs $\downarrow$ (G) \\
\midrule
No accel. & 235.65 & 1.00 & 804.26 \\
ToMe & 144.31 & 1.63 & 704.87 \\
ToDo & 145.50 & 1.62 & 717.44 \\
BiGain$_{\text{TM}}$ & 142.88 & 1.65 & 704.99 \\
BiGain$_{\text{TD}}$ & 150.30 & 1.57 & 717.44 \\
L-GTM(enc) + IE-KVD(dec) & 147.43 & 1.60 & 716.23 \\
IE-KVD(enc) + L-GTM(dec) & 146.39 & 1.61 & 712.49 \\
\bottomrule
\end{tabular}
}
\end{table}

\section{Use of Large Language Models}

We used an LLM to help solely polish the writing of the paper, while all ideas and experiments are conceived and carried out entirely by the authors.

\end{document}